\documentclass[twoside]{article}
\usepackage{amsfonts,amssymb,amsbsy,textcomp,marvosym,picins,amsmath,caption,threeparttable,amsthm,subfigure}
\usepackage{eurosym,mathrsfs,fancyhdr,CJK,multicol,graphics,indentfirst,color,bm,upgreek,booktabs,graphicx}
\usepackage{cite}
\usepackage{multirow}
\usepackage{booktabs}  
\usepackage{threeparttable}
\usepackage{pifont}
\usepackage{url}
\def\footnoterule{\kern 1mm \hrule width 10cm \kern 2mm}

\allowdisplaybreaks
\catcode`@=11
\def\title#1{\vspace{3mm}\begin{flushleft}\vglue-.1cm\Large\bf\boldmath\protect\baselineskip=18pt plus.2pt minus.1pt #1
\end{flushleft}\vspace{1mm} }
\def\author#1{\begin{flushleft}\normalsize #1\end{flushleft}\vspace*{-4pt} \vspace{3mm}}
\def\address#1#2{\begin{flushleft}\vglue-.35cm${}^{#1}$\small\it #2\vglue-.35cm\end{flushleft}\vspace{-2mm}\par}

\catcode`@=11
\def\section{\@startsection{section}{1}{\z@}%
 {-3ex \@plus -.3ex \@minus -.2ex}%
 {2.2ex \@plus.2ex}%
{\normalfont\normalsize\protect\baselineskip=14.5pt plus.2pt minus.2pt\bfseries}}
\def\subsection{\@startsection{subsection}{2}{\z@}%
 {-3ex\@plus -.2ex \@minus -.2ex}%
 {2ex \@plus.2ex}%
{\normalfont\normalsize\protect\baselineskip=12.5pt plus.2pt minus.2pt\bfseries}}
\def\subsubsection{\@startsection{subsubsection}{3}{\z@}%
 {-2.2ex\@plus -.21ex \@minus -.2ex}%
 {1.4ex \@plus.2ex}
{\normalfont\normalsize\protect\baselineskip=12pt plus.2pt minus.2pt\sl}}

\newcommand{\tabincell}[2]{\begin{tabular}{@{}#1@{}}#2\end{tabular}}

\makeatletter
\newenvironment{tablehere}
{\def\@captype{table}}
{}
\makeatother

\begin{document}
\begin{CJK*}{GBK}{song}
\vspace*{2mm}

\title{FDNet: A Deep Learning Approach with Two Parallel Cross Encoding Pathways for Precipitation Nowcasting}

\author{Bi-Ying Yan$^{1,2}$, \emph{Member}, \emph{CCF}, Chao Yang$^{3,*}$, \emph{Senior Member}, \emph{CCF}, \emph{Member}, \emph{ACM}, \emph{IEEE}, Feng Chen$^{2,4}$, Kohei Takeda$^{5}$ and Changjun Wang$^{6}$}

\address{1}{University of Chinese Academy of Sciences, Beijing 100049, China}
\address{2}{Institute of Software, Chinese Academy of Sciences, Beijing 100190, China}
\address{3}{School of Mathematical Sciences, Peking University, Beijing 100871, China}
\address{4}{Guiyang Academy of Information Technology, Guiyang 550081, China}
\address{5}{NTT DATA Corporation, Tokyo 163-8001, Japan}
\address{6}{NTT DATA Institute of Management Consulting Inc., Tokyo 163-8001, Japan}

\vspace{2mm}

\noindent E-mail: biying@iscas.ac.cn; chao\_yang@pku.edu.cn; chenfeng@iscas.ac.cn; Kohei.Takeda@nttdata.com; wanvc@nttdata-strategy.com\\[-1mm]


\let\thefootnote\relax\footnotetext{{}\\[-4mm]
\indent\ This work was supported in part by the National Key Research and Development Program of China under Grant
No.2018YFC0831500, the Beijing Natural Science Foundation under Grant No.JQ18001, and the Beijing Academy of Artificial Intelligence. \\[.5mm]
\indent\ $^*$Corresponding Author
}

\noindent {\small\bf Abstract} \quad  {\small {With the goal of predicting the future rainfall intensity in a local region over a relatively short period time, precipitation nowcasting has been a long-time scientific challenge with great social and economic impact. The radar echo extrapolation approaches for precipitation nowcasting take radar echo images as input, aiming to generate future radar echo images by learning from the historical images. To effectively handle complex and high non-stationary evolution of radar echoes, we propose to decompose the movement into optical flow field motion and morphologic deformation. Following this idea, we introduce Flow-Deformation Network (FDNet), a neural network that models flow and deformation in two parallel cross pathways. The flow encoder captures the optical flow field motion between consecutive images and the deformation encoder distinguishes the change of shape from the translational motion of radar echoes. We evaluate the proposed network architecture on two real-world radar echo datasets. Our model achieves state-of-the-art prediction results compared with recent approaches. To the best of our knowledge, this is the first network architecture with flow and deformation separation to model the evolution of radar echoes for precipitation nowcasting. We believe that the general idea of this work could not only inspire much more effective approaches but also be applied to other similar spatiotemporal prediction tasks.}}

\vspace*{3mm}


\vspace*{4mm}

\end{CJK*}
\baselineskip=18pt plus.2pt minus.2pt
\parskip=0pt plus.2pt minus0.2pt
\begin{multicols}{2}

\section{Introduction}

Precipitation nowcasting is the task of providing precise and timely prediction of rainfall intensity in a local region over a very short range (e.g., 0-6 hours) based on radar echo maps, rain gauge, and other observation data. Accurate prediction plays a vital role in our daily life as well as many public safety scenarios such as road condition alarms and flight schedules. Since the accuracy and timeliness are highly desired, compared with other traditional forecasting tasks like weekly average temperature prediction, precipitation nowcasting has become a more challenging mission in the field of weather forecasting. Of particular note is the forecasting of heavier rainfall, which occurs less often but has a higher real-world impact. Forecasting has become extremely difficult because of the low frequency of heavier rainfall events.

Existing methods for precipitation nowcasting include numerical weather prediction (NWP) based methods and radar echo extrapolation based methods~\cite{1}. NWP is one of the most successful approaches to conducting medium- and long-range (up to 6 days) weather prediction~\cite{1,2,10}. The core of NWP is the complex and meticulous simulation of the physical equations in the atmosphere model. These physical equations are coincident with our current beliefs about the dynamical behaviors of the atmosphere. The NWP-based approach has high computing complexity and its accuracy strongly relies on the initial conditions, which could make them lose competitiveness to the extrapolation-based methods in a short period forecast.

Recently, radar echo extrapolation-based methods have been noticed and widely adopted~\cite{2,3} these years. These algorithms rely on the extrapolation of observations by ground-based radars via optical flow techniques or neural network models. Optical flow-based~\cite{3,4,5} methods, as typical extrapolation-based methods, have drawn increasingly more attention, owing to their fast speeds and high accuracies. The approach is conducted in two stages by the extrapolation of radar observations. Firstly, the wind is estimated by comparing two or more precipitation fields as seen by radar based on the optical flow estimation algorithms developed in computer vision. Secondly, the precipitation field is moved along the estimated directions of the wind. However, the success of these optical flow-based methods is limited because the flow estimation step and the radar echo extrapolation step are separated and it is no easy to determine the most appropriate model parameters.

Neural network models consider precipitation nowcasting as a real application of spatiotemporal predictive learning which generate images conditioned on given consecutive frames and have shown their advantages~\cite{6,7,8} in many real-world datasets. In spatiotemporal predictive learning, there are two crucial aspects: spatial correlations and temporal dynamics, and previous studies mainly focus on how to model these two sides in a unified architecture. The Convolutional LSTM (ConvLSTM)~\cite{6} architecture naturally considers these two aspects jointly by plugging the convolutional operations in recurrent connections. Predictive RNN (PredRNN)~\cite{8} extends ConvLSTM and involves a new Spatiotemporal LSTM (ST-LSTM) unit aiming to memorize both spatial appearances and temporal variations in a unified memory pool. Just the opposite of PredRNN, MCNet~\cite{9} captures the spatial layout of an image and the corresponding temporal dynamics independently.

\begin{figure*}[!htb]
\centering
  \subfigure[]{
      \includegraphics[width=0.5\textwidth]{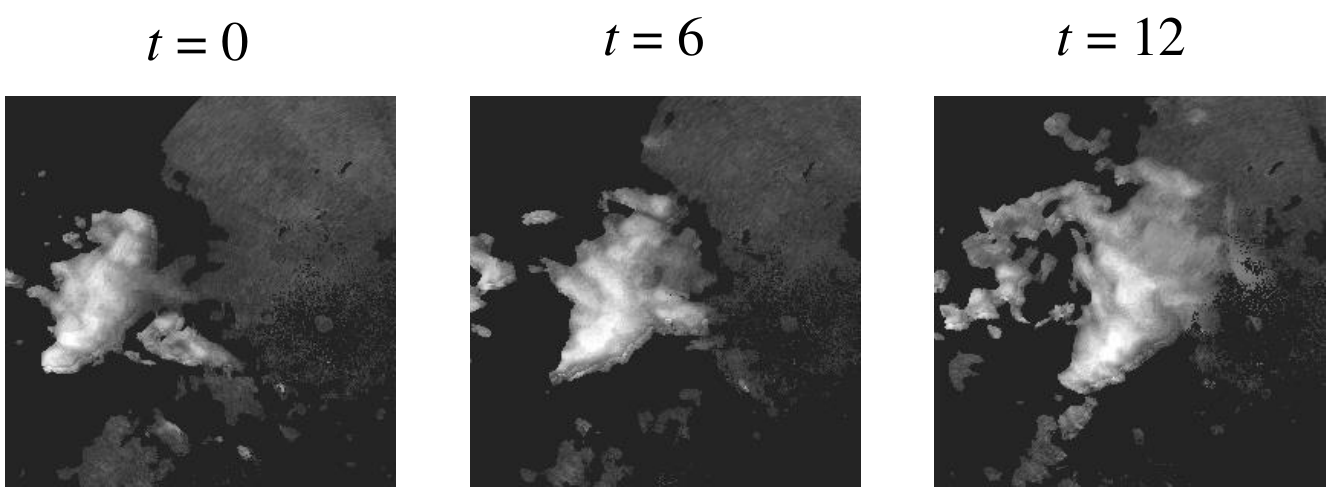}}
  \\
  \subfigure[]{
      \includegraphics[width=0.5\textwidth]{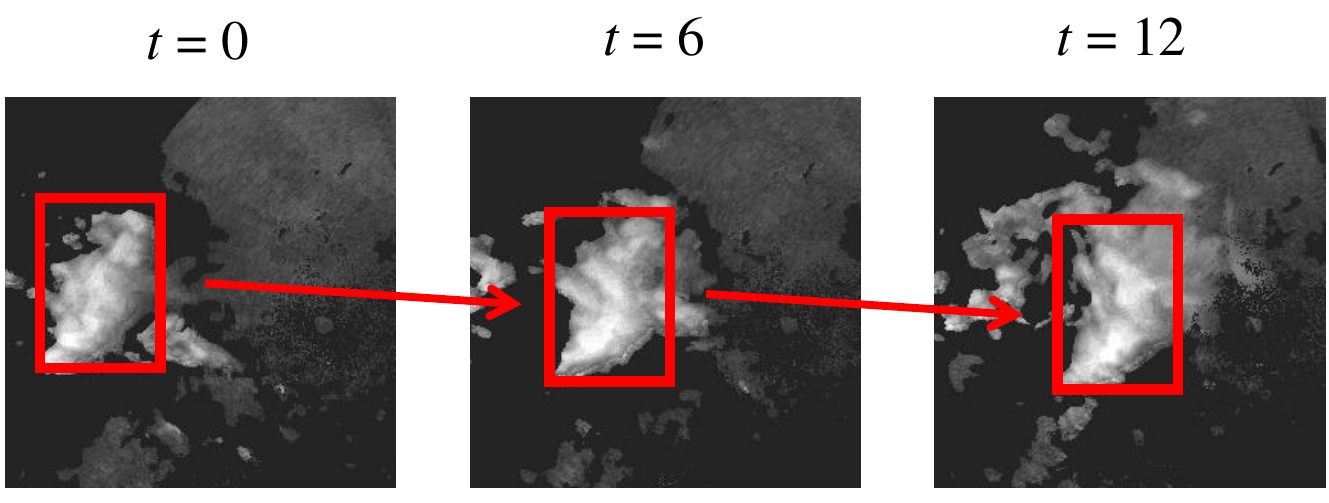}}
  \\
  \subfigure[]{
      \includegraphics[width=0.5\textwidth]{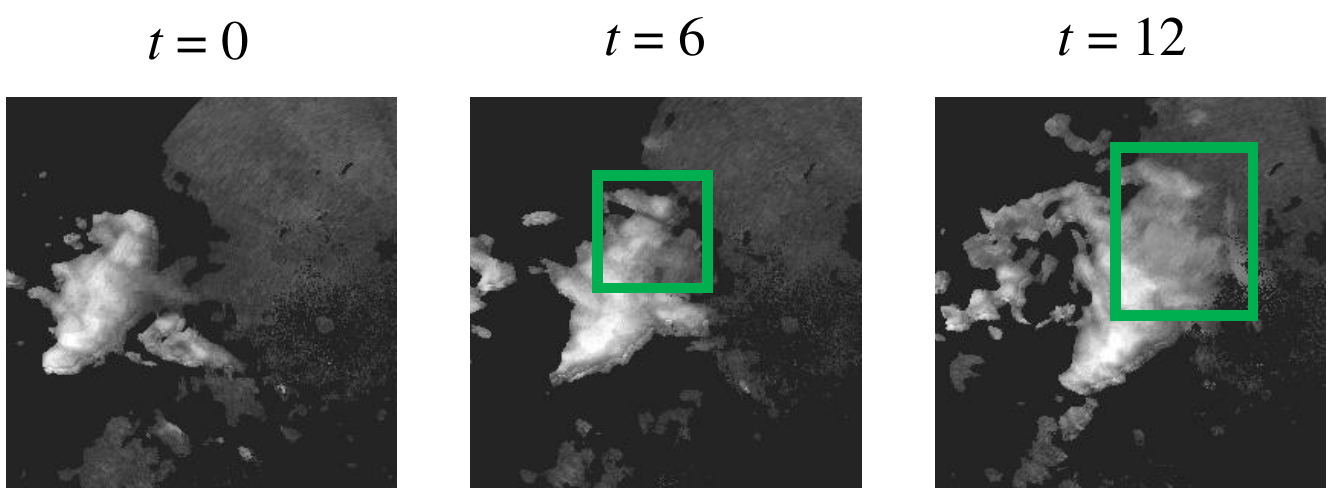}}
  \\
  \subfigure[]{
      \includegraphics[width=0.5\textwidth]{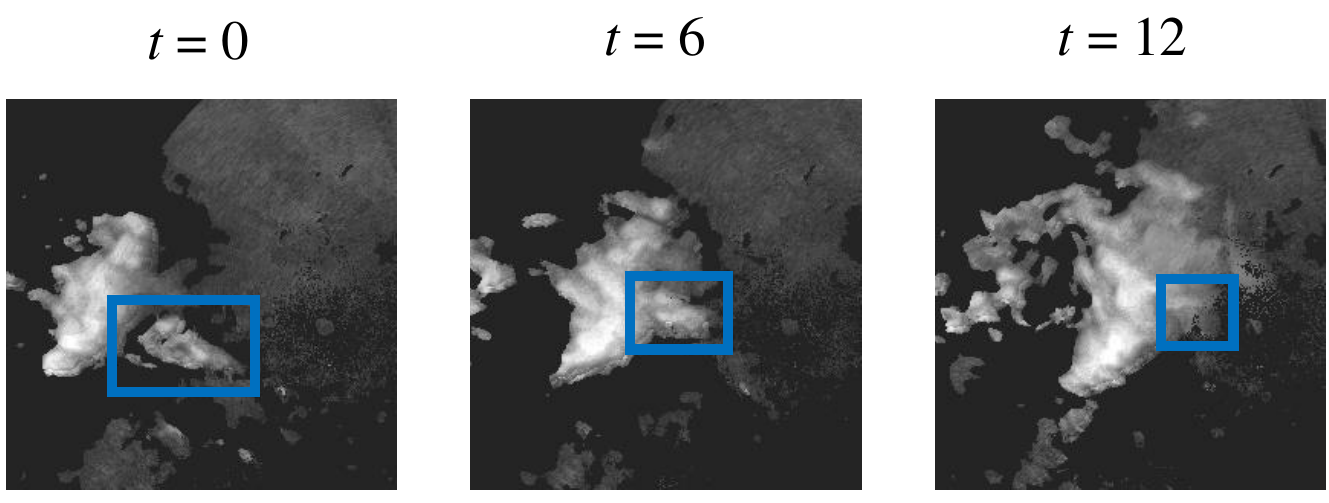}}
\vspace{2mm}
\caption{An example of 12 consecutive radar echo maps to illustrate the optical flow field motion and morphologic deformation  of radar echoes. (a) Visualizations of a typical radar echo sequence. (b) Translational motion of the main body of radar echoes (in red boxes). (c) The accumulation and deformation of the radar echoes (in green boxes). (d) The radar echoes in blue boxes dissipate and blend into the main body.}
\label{fig:intro}
\end{figure*}

In general, the evolution of radar echoes can be resolved into optical flow field motion and morphologic deformation. As shown in Fig.\ref{fig:intro}, the pixels are moving from left to right as a whole. But zooming in to the detail, the accumulation, deformation, and dissipation of the radar echoes are happening in every region at every moment. Therefore, it is difficult to capture all these fine-grained evolution patterns with a uniform encoding structure. 

\vspace{1mm}

Unfortunately, previous work only considers these two aspects of evolution together, which may lose sight of fine-grained variation. Since atmospheric motion is a complex physical process and the shapes of radar echoes may expand, contract or change rapidly, modeling morphologic deformation is significant for the prediction of radar echo maps. Motivated by this, we present a new prediction architecture called Flow Deformation Network (FDNet), which considers optical flow field motion and morphologic deformation separately, aiming to capture the fine-grained evolution of radar echoes. This model separates the information streams (position and shape) into different encoding pathways. It applys a neural optical flow estimation method on the position stream to extract the spatial coherence motion and a differencing operation on the shape stream to capture the fine-grained spatial deformation. The proposed model is evaluated on two real-world radar echo datasets, and the results show that it outperforms previous approaches, especially for longer future time steps.

The rest of the paper is organized as follows. A brief review of related work is given in Section 2 and
some preliminaries are introduced in Section 3. The detailed structure of the proposed network is described in Section 4. 
Implementation details and experimental results on challenging benchmarks are illustrated in Section 5. The paper is concluded in Section 6.

\section{Related Work}
The precipitation nowcasting task can be formulated as a spatiotemporal prediction problem in which both the input and the prediction target are spatiotemporal sequences. Convolutional neural networks (CNNs)~\cite{11} and recurrent neural networks (RNNs)~\cite{12} have been widely used for learning spatial correlations and temporal dependencies from a spatiotemporal sequence. Naturally, CNNs are applied to extract spatial features. As for temporal dynamics, the existing architecture can be generally divided into two types: (1) CNNs-based models which treat the spatiotemporal sequence as multiple channels~\cite{13,14} or the depth dimension of the image~\cite{15} and apply CNNs to capture the spatial features; (2) RNNs-based models which use RNNs to learn the variations over time~\cite{6,7,8,9}~\cite{16,17,18,19}.

ConvLSTM~\cite{6} firstly integrates CNN and RNN together by replacing the fully connection with convolutional structures in both the input-to-state and state-to-state transitions in long short-term memory recurrent neural network (LSTM). The proposal of ConvLSTM has become a milestone in the field of spatiotemporal prediction and most of the subsequent approaches are built upon ConvLSTM. Our model also employs ConvLSTM to capture the spatiotemporal correlations.

From the perspective of state transition functions, dynamic filter network (DFN)~\cite{20} extends ConvLSTM~\cite{6} by generating the filters between state transitions dynamically conditioned on input data. TrajGRU~\cite{7} actively learns the location-variant structure for recurrent connections which performs better on rotation motion. MIM~\cite{19} involves stacked multiple LSTM-similar blocks to replace the saturating forget gate in the LSTM unit and exploits the differential signals between adjacent recurrent states to model the non-stationary process. In the aspect of encoder-forecaster architecture, PredRNN~\cite{8} adds a new memory (called spatiotemporal memory) cell in each LSTM unit which is allowed to zigzag through all RNN states across different RNN layers, aiming to extract and memorize spatial and temporal representations simultaneously. 

In addition, E3D-LSTM~\cite{30} integrates 3D-convolution into RNNs to make local perceptrons of RNNs motion-aware. Conv-TT-LSTM~\cite{28} involves a higher-order convolution LSTM with a tensor-train module that combines convolutional features across time to learn long-term forecasting. Self-Attention ConvLSTM~\cite{31} introduces the self-attention mechanism into ConvLSTM to extract spatial features with both global and local dependencies. There are also efforts that extend the training data from 2D radar images to 3D images, while adapting the above ConvRNN models for the 3D-radar-extrapolation problem~\cite{32}.

However, all those above work considers the motion and deformation together, which may lose sight of fine-grained evolutions. To the best of our knowledge, only MCNet~\cite{9} decomposes the motion and content which independently captures the spatial layout and the corresponding temporal dynamics, but it does not consider the variation of the content such as accumulation, deformation, or dissipation. By contrast, our model is characterized to learn not only the tendency of global motion but also the variation of local deformation.
\section{Preliminaries}
\subsection{Structured Sequence Predictive Learning}
Sequence predictive learning is the problem of predicting the most likely future length-$K$ sequence given the previous $J$ observations:
\begin{equation}\nonumber
\begin{aligned}
\hat{\bm{x}}_{t\!+\!1}\!,...,\!\hat{\bm{x}}_{t\!+\!K}\! =\! \mathop{argmax}\limits_{\bm{x}_{t\!+\!1},...,\bm{x}_{t\!+\!K}}\!p(\bm{x}_{t+1}\!,...,\!\bm{x}_{t\!+\!K}|\bm{x}_{t\!-\!J\!+\!1},...,\bm{x}_{t}),
\end{aligned}
\end{equation}
where $\bm{x}_{t} \in D$ is an observation at time $t$ and $D$ denotes the domain of the observed features. The structured sequence is a type of special sequences where features of the observations $\bm{x}_{t}$ are not independent but linked by pairwise or spatial relationships. Such structures may be regular grid-structured like the 2D radar echo map or graph-structured like the traffic network.

In this paper, we mainly focus on the regular grid-structured sequence predictive learning problem. $\bm{x}_{t}$ can be viewed as signals on an $M \times N$ grid which consists of $M$ rows and $N$ columns. Inside each cell in the grid, $P$ measurements are varying over time. Thus, $\bm{x}_{t}$ can be represented by a tensor $\mathcal{X} \in \mathbb{R}^{P \times M \times N}$. For precipitation nowcasting, $\bm{x}_{t}$ is a 2D radar echo map.
\subsection{Convolutional LSTM} \label{section:convlstm}
Convolutional LSTM (ConvLSTM)~\cite{6} is a popular model for regular grid-structured sequences, which explicitly encodes the structured information into tensors by replacing the multiplications of dense matrices in classical LSTM with convolutions. The main equations of ConvLSTM are shown as follows:
\begin{equation}\nonumber
\begin{aligned}
\bm{g}_t &= {\rm tanh}(\bm{W}_{xg} * \bm{\mathcal{X}}_t + \bm{W}_{hg} * \bm{\mathcal{H}}_{t-1} + \bm{b}_g),\\
\bm{i}_t &= \mathop{\sigma}(\bm{W}_{xi} * \bm{\mathcal{X}}_t + \bm{W}_{hi} * \bm{\mathcal{H}}_{t-1} + \bm{W}_{ci} \odot \bm{\mathcal{C}}_{t-1} + \bm{b}_f),\\
\bm{f}_t &= \mathop{\sigma}(\bm{W}_{xf} * \bm{\mathcal{X}}_t + \bm{W}_{hf} * \bm{\mathcal{H}}_{t-1} + \bm{W}_{cf} \odot \bm{\mathcal{C}}_{t-1} + \bm{b}_f),\\
\bm{\mathcal{C}}_t &= \bm{f}_t \odot \bm{\mathcal{C}}_{t-1} + \bm{i}_t \odot \bm{g}_t,\\
\bm{o}_t &= \mathop{\sigma}(\bm{W}_{xo} * \bm{\mathcal{X}}_t + \bm{W}_{ho} * \bm{\mathcal{H}}_{t-1} + \bm{W}_{co} \odot \bm{\mathcal{C}}_{t} + \bm{b}_o),\\
\bm{\mathcal{H}}_t &= \bm{o}_t \odot {\rm tanh}(\bm{\mathcal{C}}_{t}),
\end{aligned}
\end{equation}
where $\sigma(\cdot)$ is the sigmoid activation function, $*$ and $\odot$ denote the convolution operator and the Hadamard product, respectively.

The ConvLSTM determines the future state of a certain cell in the grid by the inputs and past states of its local neighbors, which is achieved by using a convolution operator in the state-to-state and input-to-state transitions. If the states are viewed as the hidden representations of moving objects, a ConvLSTM with a larger transitional kernel should be able to capture faster motions while one with a smaller kernel can capture slower motions~\cite{6}. ConvLSTM has been adopted as a building block in many complex structures~\cite{8,9,19} since its proposal. We also employ ConvLSTM as an elementary building block in our proposed FDNet architecture.

\vspace{1mm}
\setcounter{figure}{1}
\begin{figure*}[!htb]
\centering
\includegraphics[width=0.95\textwidth]{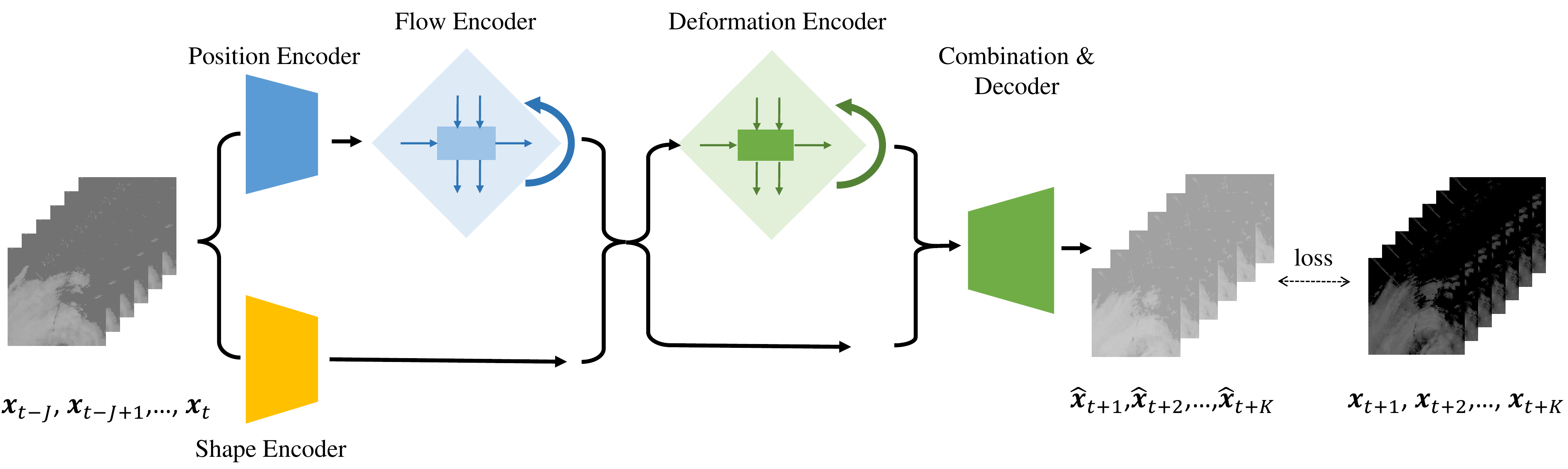}
\caption{The architecture of FDNet. FDNet is a sequence-to-sequence trainable network architecture with flow and deformation separation to model the spatio-temporal dynamics for pixel-level future prediction in radar echo maps.}
\label{fig:architecture}
\end{figure*}

\section{Model Architecture of FDNet}
We take an end-to-end learning approach to predict future radar echo maps: given a dataset consisting of the previous $J$ radar echo map sequences $\bm{x}_{t-J},\bm{x}_{t-J+1},...,\bm{x}_{t}$ and the ground truth future length-$K$ radar echo map sequences $\bm{x}_{t+1},\bm{x}_{t+2},...,\bm{x}_{t+K}$, we train a network to predict the future length-$K$ sequence directly from the previous length-$J$ sequence. The overall architecture of the proposed Flow-Deformation Network (abbreviated as FDNet) is described in Fig.\ref{fig:architecture}.

FDNet is comprised of five components: position encoder, shape encoder, flow encoder, deformation encoder, and combination \& decoder. The position encoder and the shape encoder extract meaningful position features and shape features of the radar echoes from a single frame, respectively. The flow encoder takes the spatial correspondence between two adjacent frames as an input, and feeds it into ConvLSTM layers to produce the hidden representation encoding the optical flow of the sequence. The deformation encoder takes the difference between the current moment shape representation and the predicted shape representation (computed from the previous shape representation), and the optical flow representation as an input, and outputs the hidden representation of the morphologic deformation. Finally, the combination \& decoder takes the outputs from the deformation encoder and the shape encoder as inputs, and combines them to produce a pixel-level prediction of the next frame. The prediction of multiple frames can be achieved by recursively performing the above procedures over $K$ time steps.
We will describe the detailed configuration of the proposed architecture as follows.

\vspace{1mm}
\setcounter{figure}{2}
\begin{figure*}[!htb]
\centering
  \subfigure[]{
    \includegraphics[width=0.20\textwidth]{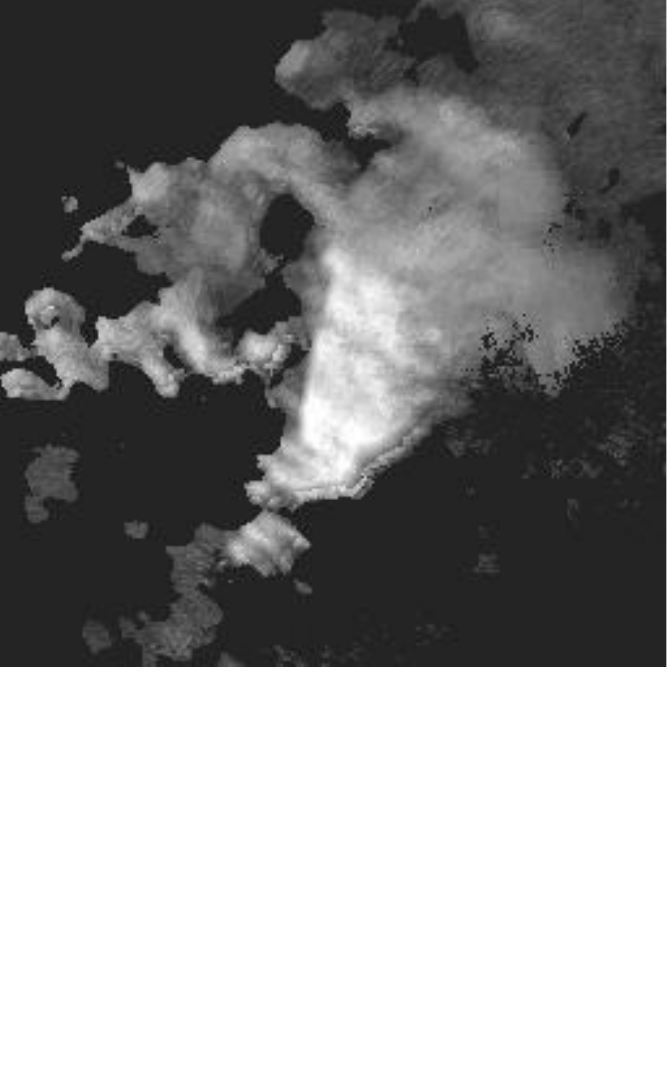}}
~
  \subfigure[]{
    \includegraphics[width=0.33\textwidth]{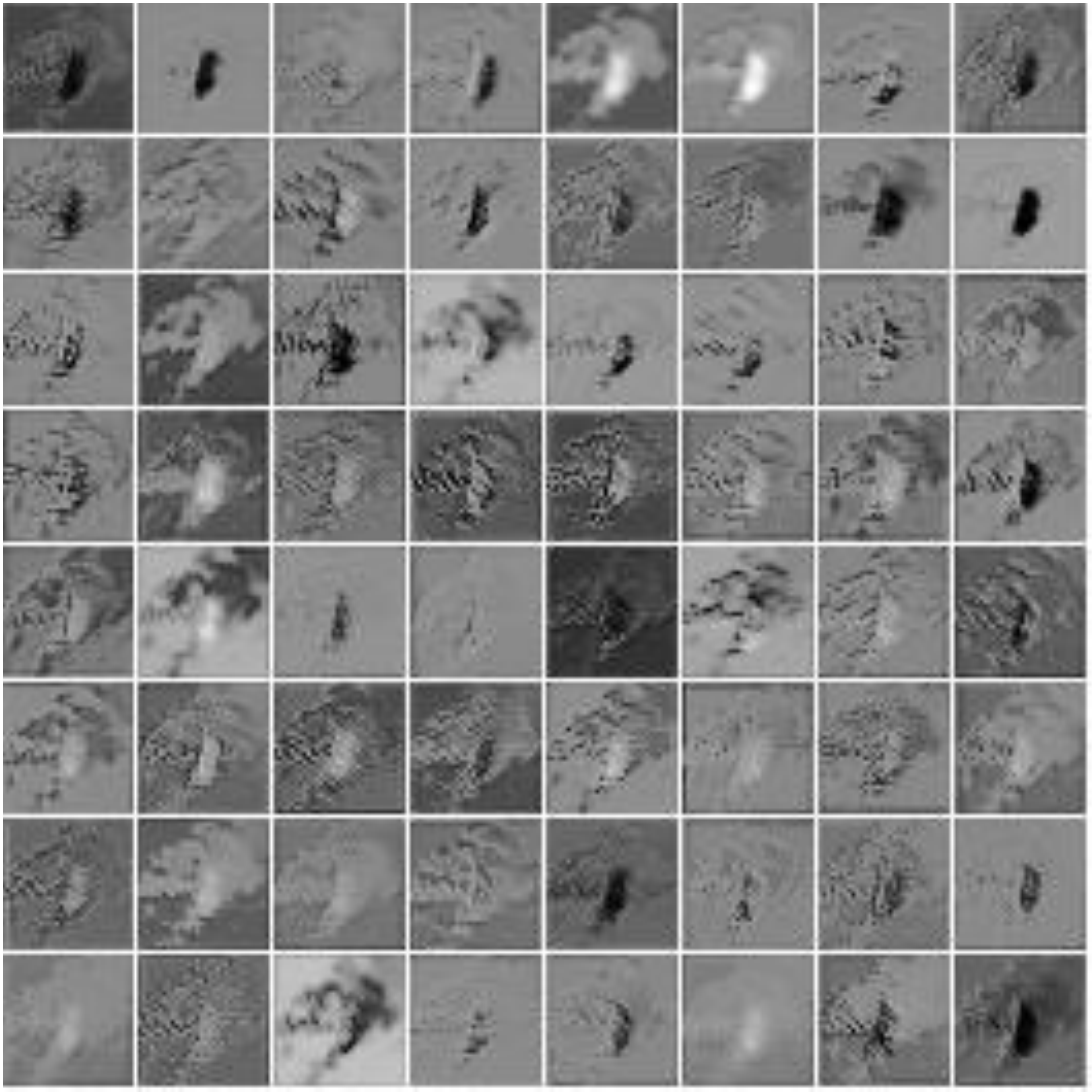}}
~
  \subfigure[]{
    \includegraphics[width=0.33\textwidth]{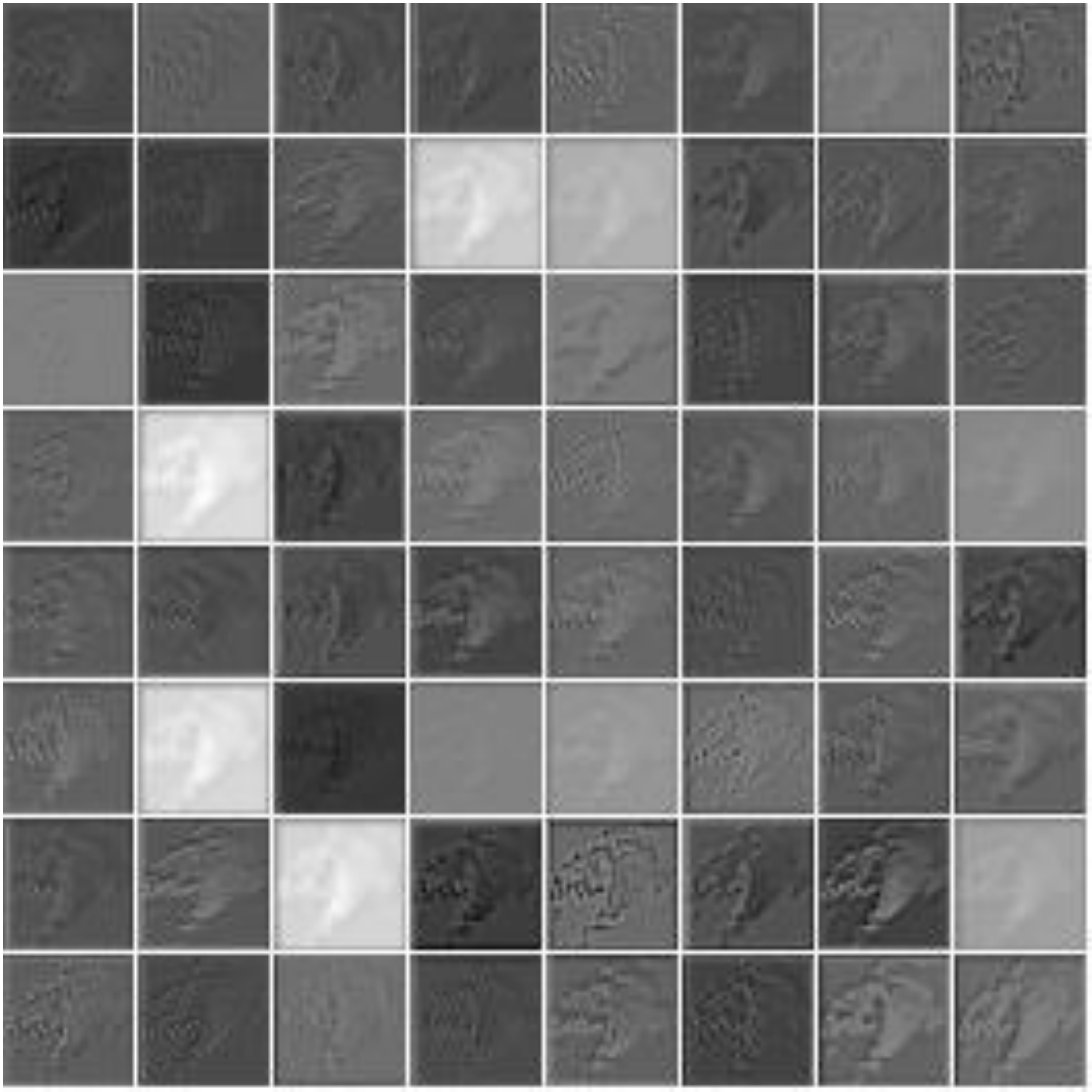}}
\caption{Visualizations of the output from the shape encoder and position encoder. (a)The original radar echo image. (b)The feature maps from the shape encoder. (c)The feature maps from the position encoder. From the output results of the position encoder, we can roughly see the outline of the radar echo. The shape encoder focuses more on pixel-level details, which is obviously different from the position encoder.}
\label{fig:shape_position}
\end{figure*}

\subsection{Position Encoder and Shape Encoder}
The position encoder extracts typical position features of pixels from a single frame in a sequence by $\bm{m}_t = {f}^{\rm pos}(\bm{x}_t)$, where $\bm{m}_t \in \mathbb{R}^{w \times h \times c}$ is the representation of the position feature of the current frame. ${f}^{\rm pos}$ is implemented by a CNN including convolution layers and activation layers. 

The shape encoder has the similar structure of the position encoder. It extracts important shape features from a single frame by  $\bm{s}_t = {f}^{\rm shape}(\bm{x}_t)$, where $\bm{s}_t \in \mathbb{R}^{w' \times h' \times c'}$ is the representation of the shape feature of the current frame. ${f}^{\rm shape}$ is also implemented by a CNN.

In order to examine whether the shape encoder and the position encoder can extract the characteristics of the shape and the position, Fig.\ref{fig:shape_position} visualizes the output feature maps from them, respectively. We can see that the model has learned to extract shape features and position features, and is working in the way as expected. The shape encoder learns the fine shape details including pixel light and shade, while the position encoder captures coarse localization and outline of the image.

\subsection{Flow Encoder}
The flow encoder aims to capture the optical flow field motion between consecutive frames without considering the deformation of the content in the frames. But how to capture the features of the flow effectively? TrajGRU~\cite{7} tackles the problem by stacking both input images together and feeds them through a rather generic convolutional neural network, allowing the network itself to decide how to process the image pair to extract the motion information. But it only roughly uses the previous time period optical flow to do the next time period transformation.

Considering that the optical flow is variable and the variation is coherent, it could be beneficial to design a strategy to extract the features of the optical flow and feed them into a time series modeling network to learn the variation. But what could be considered as the hidden representation of the optical flow? It is natural to take the correlation between two consecutive images directly as the hidden feature of the optical flow. However, how would the network find this correlation?

\vspace{1mm}
\setcounter{figure}{3}
\begin{figure*}[!htb]
\centering
  \subfigure[]{
    \includegraphics[width=0.44\textwidth]{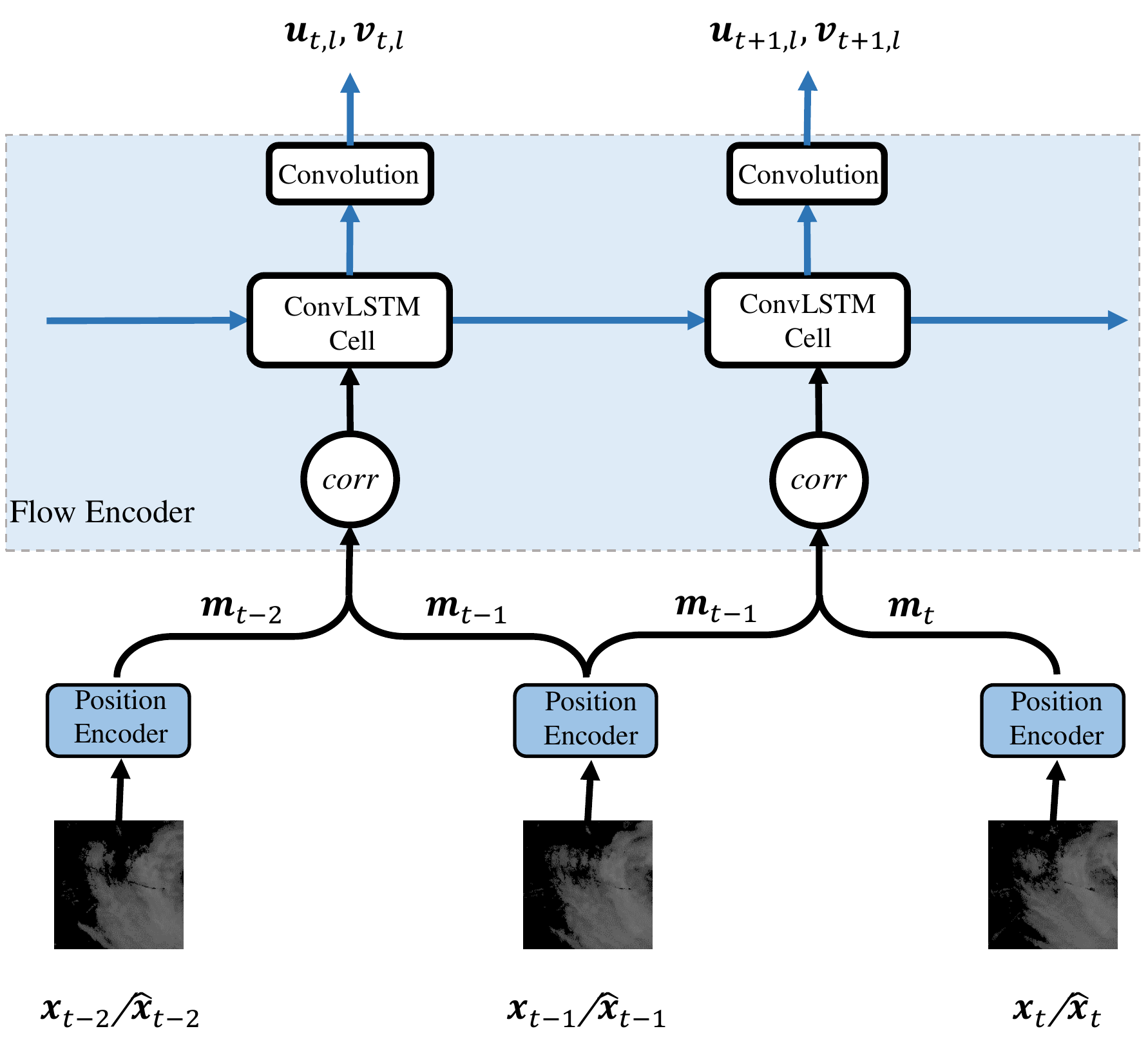}}
  \subfigure[]{
    \includegraphics[width=0.52\textwidth]{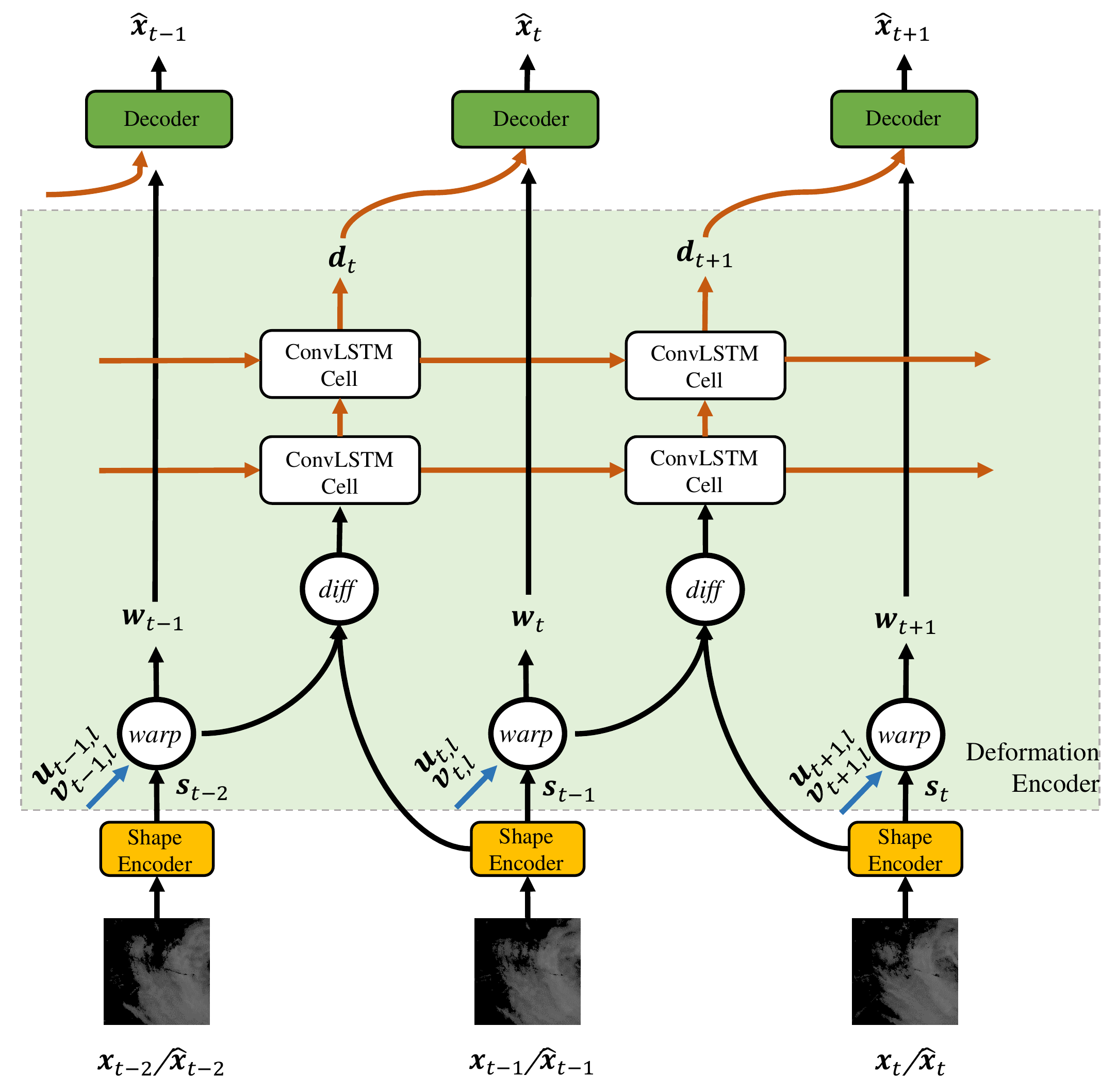}}
  \caption{The detail of FDNet. (a) The position encoder and the flow encoder. Blue arraws carry optical flow field motion information. (b) The shape encoder, deformation encoder and combination \& decoder. Brown arrows carry morphologic deformation information. The input can be either the ground truth frame for input sequence, or the generated frame at previous time step. One frame is generated at each time step as the output.}
\label{fig:detail}
\end{figure*}
\baselineskip=18pt plus.2pt minus.2pt
\parskip=0pt plus.2pt minus0.2pt

We take a similar ``\texttt{corr}" function as that in~\cite{21} to perform multiplicative patch comparisons between two feature maps. Let $\bm{m}_{t-1} \in \mathbb{R}^{w \times h \times c}$, $\bm{m}_t \in \mathbb{R}^{w \times h \times c}$ be multi-channel feature maps in time step $t-1$ and time step $t$, respectively, and $w$, $h$ and $c$ are their width, height  and the number of channels respectively. Given a maximum displacement $d$, for each location $p_{x,y}$ in the first feature map $\bm{m}_{t-1}$, the \texttt{corr} function will compute the correlations between the feature of $p_{x,y}$ and every point $q$ in the region $[x-d,y-d] \times [x+d,y+d]$ in the second feature map $\bm{m}_t$. The ``correlation'' between point $p_{x_1,y_1}$ in the first feature map and point $q_{x_2,y_2}$ in the second feature map is defined as the dot product of feature vectors in point $p_{x_1,y_1}$ and point $q_{x_2,y_2}$:
\begin{equation}\nonumber
\begin{aligned}
corr(p_{x_1,y_1},q_{x_2,y_2}) = \bm{m}_{t-1}[x_1][y_1] \cdot \bm{m}_t[x_2][y_2].
\end{aligned}
\end{equation}

Computing correlations among all patch combinations involves $w^2 \cdot h^2 \cdot d^2$ computations, which is time consuming. For computational reasons we limit the maximum displacement $d \approx w/3$. Specifically, if the size of feature map is $64 \times 64$, we set $d=21$, and if the size of feature map is $32 \times 32$, we set $d=11$. Besides, we introduce striding in both feature maps, and in our experiment we set stride to 2.

The result produced by the \texttt{corr} function is three dimensional: for every combination of two 2D positions we obtain a correlation value, and there are $(2d+1)^2$ combinations, and thus we obtain a final output of size $w \times h \times (2d+1)^2$.

The output of the \texttt{corr} function will be fed into a ConvLSTM (see subsection~\ref{section:convlstm}) layer to encode the correlations. The advantage of such a structure is that we could learn the variation of optical flow by learning the parameters of the ConvLSTM.

After that, we take a one-hidden-layer convolutional neural network to extract the optical flow. TrajGRU~\cite{7} claims that motion patterns have different neighborhood sets for different locations and therefore the ``optical flow" should be not just one but a set. In our experiments, we tested both single optical flow and multiple optical flows. But the results show that the single optical flow approach achieves a better performance. 

As described in Fig.\ref{fig:detail}(a), the flow encoder takes the position encoder $\bm{m}_{t-2}$ and $\bm{m}_{t-1}$ at time step $t-2$ and time step $t-1$ as inputs, respectively, and produces the hidden representations of optical flow at time step $t$, which is denoted by $\bm{u}_{t,l},\bm{v}_{t,l}$. Fig.\ref{fig:flow}(a) visualizes the ``flow'' information that the flow encoder outputs. We can see that the network has learned reasonable optical flow field motion information.

\vspace{1mm}
\setcounter{figure}{4}
\begin{figure*}[!htb]
\centering
  \subfigure[]{
    \includegraphics[width=0.99\textwidth]{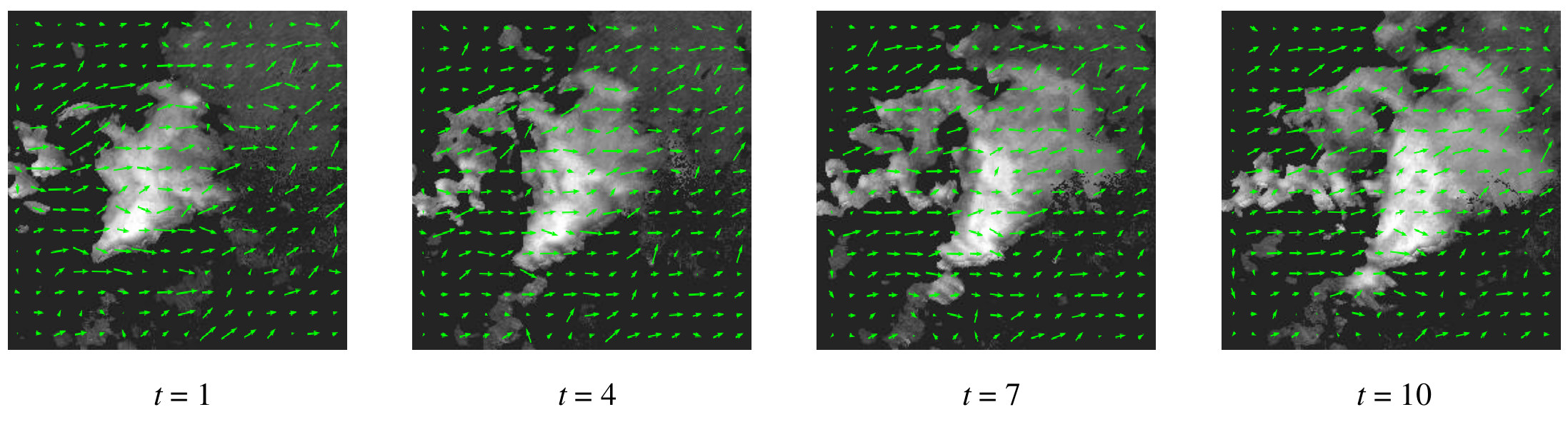}}
  \subfigure[]{
    \includegraphics[width=0.99\textwidth]{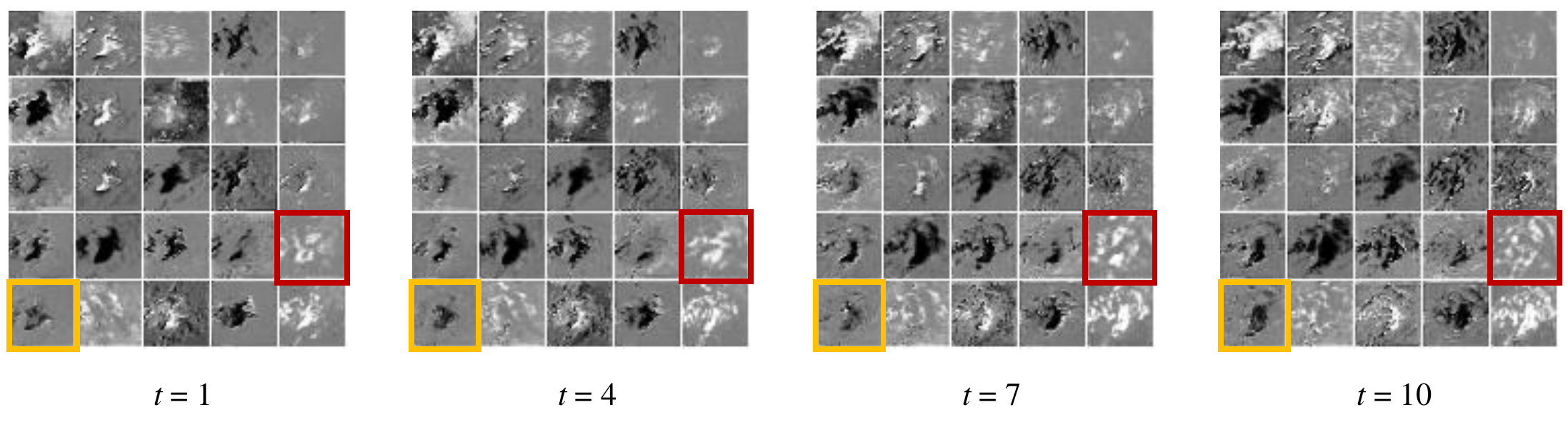}}
\caption{The visualization of the output from the flow encoder and the deformation encoder. (a) The visualization of the optical flow field motion that the network has learned. From left to right, we select the radar echo images at $t$=1, 4, 7, and 10. It can be seen that the optical flow field information learned by the model is basically consistent with the actual motion of the radar echoes. (b) The visualization of output feature maps of the deformation encoder. We select the corresponding feature maps with the same time step as that in (a).}
\label{fig:flow}
\end{figure*}

\subsection{Deformation Encoder}
The evolution of radar echo map sequence is a complex process. The shapes of radar echo maps may expand, contract, or change rapidly due to the complex atmospheric environment, which is quite different from that of other video sequences like moving digits~\cite{6} and KTH actions~\cite{8}. Therefore, it is necessary to distinguish the changes of shapes from the translational motions of radar echoes, and this idea inspires the design of the deformation encoder.

But what could be considered as the hidden representation of the deformation? In our study, the transformation of two adjacent radar echo maps is divided into two aspects: the optical flow fields movement and the shape deformation. Therefore the shape feature of the former image is conducted by a ``\texttt{warp}" function same with that in ~\cite{7} to let the precipitation field move along the estimated optical flow that the flow encoder outputs, and then the deformation is computed by the difference between the result and the shape feature of the latter image.

Let $\bm{s}_{t-1}$ be the shape feature of the image at $t-1$ time step, and $\bm{u}_{t,l},\bm{v}_{t,l}$ be $l$ optical flows that the flow encoder outputs. ${warp}(\bm{s}_{t-1},\bm{u}_{t,l},\bm{v}_{t,l})$ selects the positions pointed out by $\bm{u}_{t,l},\bm{v}_{t,l}$ from $\bm{s}_{t-1}$ via the bilinear sampling kernel~\cite{22,23}. If we denote $\bm{\mathcal{M}} = { warp}(\bm{\mathcal{I}},\bm{U},\bm{V})$ where $\bm{\mathcal{M,I}} \in \mathbb{R}^{C \times H \times W}$ and $\bm{U},\bm{V} \in \mathbb{R}^{H \times W}$, we have:
\begin{equation}\nonumber
\begin{aligned}
\mathcal{M}_{c,i,j} &= \sum_{m=1}^H\sum_{n=1}^W{\mathcal{I}}_{c,m,n} \times {\rm max}(0,1-|i+V_{i,j}-m|)\\
&\times {\rm max}(0,1-|i+U_{i,j}-n|).
\end{aligned}
\end{equation}

We denote the output of \texttt{warp} function on $\bm{s}_{t-1}$ as $\bm{w}_t$. Intuitively, $\bm{w}_t$ can be viewed as the inferred shape feature at time step $t$, which is generated by transforming $\bm{s}_{t-1}$ using the predicted optical flow decoded $\bm{u}_{t,l},\bm{v}_{t,l}$. We deliver $\bm{w}_t$ and shape features of the image at time step $t$ to a ``\texttt{diff}" function, which is defined as element-wise subtraction. The output of the \texttt{diff} function will be fed into a stack of ConvLSTM layers to encode the deformation, denoted as $\bm{d}_{t+1}$, as described in Fig.\ref{fig:detail}(b). We use the hidden features out from the ConvLSTM layers rather than the \texttt{diff} function as the encoder of the deformation, aiming to learn the evolution of it over time.

The output feature maps of the deformation encoder are visualized in Fig.\ref{fig:flow}(b). It can be observed that the deformation encoder captures radar echo changes at different hierarchies. For example, the feature maps in orange boxes pay more attention to the darkest pixels, while the feature maps in red boxes are sensitive to the pixel changes in multiple scattered areas.

\subsection{Combination \& Decoder}
The outputs from the above deformation encoder, $\bm{w}_t$ and $\bm{d}_t$, encode a high-level representation of inferred image after optical flow field movement and shape deformation, respectively. Given these representations, the goal of the decoder is to generate a pixel-level prediction of the next frame $\hat{\bm{x}}_{t+1} \in \mathbb{R}^{w \times h \times c}$. For this purpose, it first combines these two representations into a unified representation by $\bm{c}_t = {g}^{\rm comb}([\bm{d}_t,\bm{w}_t])$, where $[\bm{d}_t, \bm{w}_t] \in \mathbb{R}^{w' \times h' \times 2c'}$ denotes the concatenation of $\bm{d}_t$ and $\bm{w}_t$ in the depth dimension, and $\bm{c}_t \in \mathbb{R}^{w' \times h' \times c'}$ denotes the combined high-level representation of optical fields motion. ${g}^{\rm comb}$ is implemented by a CNN layer.

Then $\bm{c}_t$ is delivered to the decoder, which places $\bm{c}_t$ back into the original pixel space by $\hat{\bm{x}}_{t+1} = {g}^{\rm dec}(\bm{c}_t)$.

We employ the deconvolution network~\cite{24} for our decoder network ${g}^{\rm dec}$, which is composed of multiple successive operations of deconvolution, rectification, and convolution.

\section{Experiments}
\subsection{Dataset Description}
We verify our model on two real-world radar echo datasets, HKO-7~\cite{7} and SRAD2020~\footnote{https://competition.huaweicloud.com/information/1000040092/circumstance?track=107}, collected by Hong Kong Observatory and Shenzhen Meteorological Bureau, respectively. Each dataset contains radar CAPPI reflectivity images recorded every 6 minutes, as detailed below.

$\bullet$ {\bfseries HKO-7}

The HKO-7 dataset contains radar echo data from 2009 to 2015 collected by Hong Kong Observatory. The radar CAPPI reflectivity images, which have resolution of 480 $\times$ 480 pixels, are taken from an altitude of 2\;km and cover an 512\;km $\times$ 512\;km area centered in Hong Kong. The raw logarithmic radar reflectivity factors are linearly transformed to pixel values via $pixel =\lfloor 255 \times \frac{dBZ + 10}{70} + 0.5 \rfloor$ and are clipped to be between 0 and 255. The dataset contains 800\;days for training, 50\;days for validation and 120\;days for testing. We use the previous 5 time steps radar echo maps to predict 20 time steps into the future, covering the next two hours. The distribution of different rainfall intensity of the data is as Table~\ref{tab:hko-statistics}.
\tabcolsep 5pt
\renewcommand\arraystretch{1.3}
\begin{tablehere}
\begin{center}
\caption{\label{tab:hko-statistics} Rain Rate Statistics in the HKO-7 Dataset}\vspace{2mm}
{\footnotesize
\begin{tabular*}{\linewidth}{ccl}
\toprule
  Rain Rate (mm/h) & Proportion(\%) & Rainfall Level \\\hline
  $0 \le x \textless 0.5$ & 89.94 & No / Hardly noticeable \\
  $0.5 \le x \textless 2$ & 4.32 & Light \\
  $2 \le x \textless 5$ & 2.51 & Light to moderate \\
  $5 \le x \textless 10$ & 1.63 & Moderate \\
  $10 \le x \textless 30$ & 1.17 & Moderate to heavy \\
  $30 \textless x $ & 0.43 & Rainstorm warning \\
\bottomrule
\end{tabular*}
\begin{tablenotes}
    \item[1] Note: $x$ means the rain rate.
\end{tablenotes}
}
\end{center}
\end{tablehere}

$\bullet$ {\bfseries SRAD2020}

The SRAD2020 dataset contains radar echo data collected by Shenzhen Meteorological Bureau in recent years. The radar CAPPI reflectivity images which have resolution of 256 $\times$ 256 pixels, are taken from an altitude of 2.5\;km and covering an 255\;km $\times$ 255\;km area. There are 20000 radar echo data cases in the SRAD2020 dataset, each of which covers four hours and is a 41-length radar echo sequence. We filter the noisy data cases which have abrupt all-zero radar echo data in a sequence. The final dataset has 15939 data cases for training, 1002 data cases for validation and 2541 data cases for testing. We predict 20 time steps into the future covering the next two hours by observing 21 time steps. The statistical distribution of radar reflectivity values in the dataset is as Table~\ref{tab:srad-statistics}.
\setcounter{table}{1}
\tabcolsep 5pt
\renewcommand\arraystretch{1.3}
\begin{tablehere}
\begin{center}
\caption{\label{tab:srad-statistics} Radar Reflectivity Statistics in the SRAD2020 Dataset}
\vspace{2mm}
{\footnotesize
\begin{tabular*}{0.9\linewidth}{ccl}
\toprule
  dBZ & Proportion(\%) & Rainfall Level \\\hline
  $0 \le x \le 20$ & 77.53 & No to light \\
  $20 \le x \textless 30$ & 11.35 & Light to moderate\\
  $30 \le x \textless 40$ & 8.55 & Moderate to heavy \\
  $40 \le x \textless 50$ & 2.37 & Heavy to rainstorm \\
  $50 \le x \le 80$ & 0.20 &\tabincell{l}{Large rainstorm to \\extraordinary rainstorm} \\
\bottomrule
\end{tabular*}
\begin{tablenotes}
    \item[1] Note: $x$ means the radar reflectivity factor.
\end{tablenotes}
}
\end{center}
\end{tablehere}

\subsection{Experiment Setting}

\subsubsection{Evaluated Algorithms}

We compare our method with five competitive precipitation nowcasting models (including ConvLSTM~\cite{6}, TrajGRU~\cite{7}, PredRNN~\cite{8}, MIM~\cite{19} and ConvTTLSTM~\cite{28}
). 
We use one layer ConvLSTM with 128 hidden states for the flow encoder and 2 layers of ConvLSTM with 128 hidden states for the deformation encoder in FDNet. 
For the ConvLSTM model and the TrajGRU model, we use a 3-layer encoding-forecasting structure with the number of hidden states for the RNNs setting to 64, 192, 192 following ~\cite{7}.
We also use a stack of 3-layer of ST-LSTM with 128 hidden states for the PredRNN model and the MIM model.
For the Conv-TT-LSTM model, we use a stack of 8-layer of Conv-TT-LSTM
with 64 hidden states for the first layer and 128 hidden states for the other layers. The order of CTTD is set to 3, the rank of CTTD is set to 8 and the time step of CTTD is set to 3 for the Conv-TT-LSTM model. 
The convolution filters inside ConvLSTMs, ST-LSTMs and Conv-TT-LSTMs are all set to $3 \times 3$. Since the ConvLSTM model and the TrajGRU model reduce the size of feature maps at higher layers, for fair comparison, we use dilated convolution at higher layers to get a larger receptive field for FDNet, PredRNN, MIM and Conv-TT-LSTM.

Considering the memory and computation factors, we conduct downsampling on the original images to reduce the resolution, and after prediction upsampling is used to restore the resolution. The 2D-CNN encoders and the 2D-CNN decoders are similar in FDNet, PredRNN, MIM and Conv-TT-LSTM. Table~\ref{Tbl:en-de} shows the details of the encoder and the decoder.
\setcounter{table}{2}
\tabcolsep 5pt
\renewcommand\arraystretch{1.3}
\begin{table*}[!ht]
\centering
\begin{threeparttable} 
\caption{\label{Tbl:en-de} Details of the Encoder and the Decoder}\vspace{-2mm}
\footnotesize
\begin{tabular*}{1\textwidth}{cccccccc}
\toprule
Module & Layer & Kernel & Stride & Padding & Output-padding & Channel I/O & Type \cr  
    \midrule
\multirow{6}{*}{encoder}&econv1&	$3\times 3$&	$2\times 2$&	$1\times 1$&	-	&1/8		&2D-Conv, GroupNorm, LeakyReLU \\
&econv2&	$3\times 3$&	$1\times 1$&	$1\times 1$&	-	&8/16		&2D-Conv, GroupNorm, LeakyReLU\\
&econv3&	$3\times 3$&    $2\times 2$&	$1\times 1$&	-	&16/32		&2D-Conv, GroupNorm, LeakyReLU\\
&econv4&	$3\times 3$&	$1\times 1$&    $1\times 1$&	-	&32/32		&2D-Conv, GroupNorm, LeakyReLU\\
&econv5&	$3\times 3$&	$2\times 2$&    $1\times 1$&	-	&32/64		&2D-Conv, GroupNorm, LeakyReLU\\
&econv6&	$3\times 3$&	$1\times 1$&	$1\times 1$&	-	&64/64		&2D-Conv, GroupNorm, LeakyReLU\\
\midrule
\multirow{6}{*}{decoder}&dconv1&	$3\times 3$&	$1\times 1$&	$1\times 1$&	1	&128/64		&Transposed 2D-Conv, GroupNorm, LeakyReLU\\
&dconv2&	$3\times 3$&	$2\times 2$&	$1\times 1$&	2	&64/32		&Transposed 2D-Conv, GroupNorm, LeakyReLU\\
&dconv3&	$3\times 3$&	$1\times 1$&	$1\times 1$&	1	&32/32		&Transposed 2D-Conv, GroupNorm, LeakyReLU\\
&dconv4&	$3\times 3$&	$2\times 2$&	$1\times 1$&	2	&32/16		&Transposed 2D-Conv, GroupNorm, LeakyReLU\\
&dconv5&	$3\times 3$&	$1\times 1$&	$1\times 1$&	1	&16/8		&Transposed 2D-Conv, GroupNorm, LeakyReLU\\
&dconv6&	$3\times 3$&	$2\times 2$&	$1\times 1$&	1	&8/1		&Transposed 2D-Conv\\
\bottomrule
\end{tabular*}
\begin{tablenotes}
    \item[1] Note: ``2D-Conv" means 2D-Convolution, ``Transposed 2D-Conv" means Transposed 2D-Convolution and ``GroupNorm" means GroupNormalization.
\end{tablenotes}
\end{threeparttable}
\end{table*}
\baselineskip=18pt plus.2pt minus.2pt
\parskip=0pt plus.2pt minus0.2pt

\subsubsection {Loss Function}
We use an objective function composed of multiple weighted losses for all models. Given the training data $D=\left\{\bm{x}_{1,...,t,...,T}^{(n)}\right\}_{n=1}^N$ where $\bm{x}^{(n)}_t \in \mathbb{R}^{W \times H}$, we use $\bm{x}_1^{(n)},...,\bm{x}_t^{(n)}$ to predict $\bm{x}_{t+1}^{(n)},...,\bm{x}_T^{(n)}$. Our model is trained to minimize the prediction loss by 
\begin{equation}\nonumber
\begin{aligned}
\mathcal{L}&=\sum_{n=1}^N\mathcal{L}_{\rm img}(\bm{x}^{(n)},\hat{\bm{x}}^{(n)})\\
&=\lambda_{\rm pixel}\mathcal{L}_{\rm pixel}(\bm{x}^{(n)},\hat{\bm{x}}^{(n)}) + \lambda_{\rm gdl}\mathcal{L}_{\rm gdl}(\bm{x}^{(n)},\hat{\bm{x}}^{(n)}).
\end{aligned}
\end{equation}

Here, $\bm{x}$ and $\hat{\bm{x}}$ are the target and predicted frames, respectively. $\lambda_{\rm pixel}$ and $\lambda_{\rm gdl}$ are hyperparameters that control the effect of each sub-loss during optimization. $\mathcal{L}_{\rm pixel}$ is designed to guide the model to match the average pixel values directly, while $\mathcal{L}_{\rm gdl}$ is expected to guide the model to match the gradients of such pixel values, to alleviate the image blurring tendency of predicted frames. In our experiments, we set $\lambda_{\rm pixel}=1$ and 
$\lambda_{\rm gdl}=1$, which means that these two aspects have the same effect.

As shown in Fig.\ref{tab:hko-statistics} and Fig.\ref{tab:srad-statistics}, the frequencies of different rainfall levels are highly imbalanced. It is known that heavier rainfalls always have much greater social impacts. Since it has been proved that training with balanced loss function is essential for good nowcasting performance of heavier rainfall~\cite{7}, we use the weighted loss function in the sub-loss $\mathcal{L}_{\rm pixel}$. Therefore we get 
\begin{equation}\nonumber
\begin{aligned}
\mathcal{L}_{\rm pixel}(\bm{y},\bm{z}) = \sum_{p}\sum_{k=1}^{T}\sum_{i,j}^{W,H}w_{k,i,j}||y_{k,i,j}-z_{k,i,j}||_p^p,
\end{aligned}
\end{equation}
where $\bm{y},\bm{z} \in \mathbb{R}^{T \times W \times H}$ are two $T$-length frame sequences, $y_{k,i,j}$ is the $(i,j)$th pixel value in the $k$th frame in $\bm{y}$, and $z_{k,i,j}$ is the $(i,j)$th pixel value in the $k$th frame in $\bm{z}$. $w_{k,i,j}$ is the weight corresponding to the $(i,j)$th pixel in the $k$th frame, relating to its rainfall intensity or radar reflectivity value.

In our experiments, we set p=1, 2, and thus we get 
\begin{equation}\nonumber
\begin{aligned}
\mathcal{L}_{\rm pixel}(\bm{y},\bm{z}) &= \sum_{k=1}^{T}\sum_{i,j}^{W,H}w_{k,i,j}(|y_{k,i,j}-z_{k,i,j}| \\
&+|y_{k,i,j}-z_{k,i,j}|^2).
\end{aligned}
\end{equation}

As for the HKO-7 dataset, if we define $r$ as the corresponding rainfall intensity of the pixel $x$, we set the weight of pixel $x$ as:
\begin{equation}
w(r) = \left\{\begin{array}{ll}
1,&r \le 2;\\
2,&2 \le r \leq 5;\\
5,&5 \le r \leq 10;\\
10,&10 \le r \leq 30;\\
30,&r \geq 30.
\end{array}
\right.
\label{equa:weight1}
\end{equation}

As for the SRAD2020 dataset, we set the weight of the pixel $x$ as:
\begin{equation}
w(x) = \left\{\begin{array}{ll}
1,&0 \le x \leq 20;\\
2,&20 \le x \leq 30;\\
5,&30 \le x \leq 40;\\
10,&40 \le x \leq 50;\\
30,&50 \le x \le 80.
\end{array}
\right.
\label{equa:weight2}
\end{equation}

However, both $\mathcal{L}_1$ loss and $\mathcal{L}_2$ loss produce blurry predictions, increasingly worse when predicting further in the future. To sharpen the image prediction, we take a gradient difference loss (GDL)~\cite{13} as a mitigation strategy. The GDL loss $\mathcal{L}_{\rm gdl}(\bm{y},\bm{z})$ is defined as:
\begin{equation}\nonumber
\begin{aligned}
\mathcal{L}_{\rm gdl}(\bm{y},\!\bm{z})\!&=\!\sum_{k=1}^{T}\!\sum_{i,j}^{W,H}(|(|y_{k,i,j}\!-\!y_{k,i-1,j}|\!-\!|z_{k,i,j}-\!z_{k,i-1,j}|)|^\lambda\\
&+ |(|y_{k,i,j}-y_{k,i,j-1}|-|z_{k,i,j}-z_{k,i,j-1}|)|^\lambda),
\end{aligned}
\end{equation}
where $\lambda \geq 1$ and $\lambda$ is an integer. Considering the training time, we only set $\lambda = 1$, which means that we only take the differencing between adjacent pixels.

We also conduct experiments without GDL loss on the SRAD2020 dataset, and the results show that the GDL loss improves the CSI score and the HSS score by 0.007 for all models on average.

\subsubsection{Optimizer and Learning Rate}

We train all compared models using Pytorch and optimize them to convergence using ADAM~\cite{25}. 
For ConvLSTM, TrajGRU, PredRNN and MIM, we set the initial learning rate to 0.001, and for FDNet, we set the initial learning rate to 0.0001. We use gradient clipping with clipping value 50 for these models. As Su~\cite{28} reports, we find that the Conv-TT-LSTM model is unstable at a high learning rate such as 0.001, but learns poorly at a low learning rate 0.0001. Therefore we use gradient clipping with learning rate 0.001 and clipping value 1 following ~\cite{28}.

We apply the scheduled sampling strategy~\cite{26} to all of the models, to gently change the training process from a fully guided scheme using the true previous data, towards a less guided scheme which mostly uses the generated data instead.

\subsubsection{Parameter Initialization}

All convolutional kernels are initialized by Xavier's normalized initializer~\cite{29}, and the initial hidden/cell states in ConvLSTM, ST-LSTM and Conv-TT-LSTM
are initialized as zeros.

\subsection{Experimental Results}
To evaluate the performance of our model, we measure the balanced mean square error (BMSE) and balanced mean average error (BMAE)~\cite{7}. BMSE is defined as:
\begin{equation}\nonumber
\begin{aligned}
BMSE = \frac{1}{N}\sum_{k=1}^{N}\sum_{i,j}^{W,H}w(x_{k,i,j})(x_{k,i,j}-\hat{x}_{k,i,j})^2,
\end{aligned}
\end{equation}
and BMAE is defined as:
\begin{equation}\nonumber
\begin{aligned}
BMAE = \frac{1}{N}\sum_{k=1}^{N}\sum_{i,j}^{W,H}w(x_{k,i,j})|x_{k,i,j}-\hat{x}_{k,i,j}|, 
\end{aligned}
\end{equation}
where $N$ is the total number of frames and $w(\cdot)$ is the weight defined in Equation (\ref{equa:weight1}) and (\ref{equa:weight2}).

We also calculate the critical success index (CSI) and Heidke skill score (HSS) for multiple thresholds that correspond to different rainfall levels. We choose to use the thresholds 20\;dBZ, 30\;dBZ, 40\;dBZ and 50\;dBZ. The CSI score is defined as: ${CSI=\frac{TP}{TP+FN+FP}}$ and the HSS score is defined as ${HSS=\frac{TP \times TN - FN \times FP}{(TP + FN)(FN+TN)+(TP+FP)(FP+TN)}}$, where $TP$ means true positive, $FP$ means false positive, $TN$ means true negative and $FN$ means false negative. A higher CSI score or a higher HSS score indicates a better prediction result.

\setcounter{table}{3}
\tabcolsep 5pt
\renewcommand\arraystretch{1.3}
\begin{table*}[!tbp]
\centering
\begin{threeparttable} 
\caption{\label{Tbl:bmse} Performance Comparisons of Different Approaches for Precipitation Nowcasting}\vspace{-2mm}
\footnotesize
\begin{tabular*}{1\textwidth}{ccccccccccc}
\toprule
   \multirow{2}{*}{Model}  & \multicolumn{5}{c}{HKO-7} & \multicolumn{5}{c}{SRAD2020} \cr
  \cmidrule(lr){2-6} \cmidrule(lr){7-11} 
    & AVG & 30min & 60min & 90min & 120min   & AVG & 30min & 60min & 90min & 120min \cr  
    \midrule
ConvLSTM~\cite{6}     & 5806.72 & 3714.66 & 5861.75 & 7609.86 & 9120.39 & 1753.22 & \textbf{1174.59} & 1801.63 & 2284.70 & 2669.52  \\ 
TrajGRU~\cite{7}      & 5818.12 & 3717.82 & 5872.09 & 7619.2  & 9170.61 & 1721.95 & 1197.60 & 1767.28 & 2273.62 & 2546.11  \\
PredRNN~\cite{8}     & 5785.6  & \textbf{3698.48} & 5865.6  & 7597.9  & 9048.29 & 1730.27 & 1195.47 & 1777.83 & 2219.29 & 2623.05  \\
MIM~\cite{19}       & 5784.17 & 3701.48 & 5854.5  & 7571.12 & 9045.05 & 1835.4  & 1292.99 & 1885.64 & 2336.53 & 2713.84  \\
Conv-TT-LSTM~\cite{28}     &  6104.29 & 4012.97 & 6096.10  & 7852.03 & 9373.00 & 1970.42  & 1312.84 &  2013.77 & 2549.43 & 2993.48  \\
\textbf{FDNet}     &  \textbf{5781.21}   &   3716.69       &   \textbf{5842.25}      &   \textbf{7564.48}       &   \textbf{9043.71}      &  \textbf{1671.1}  &  1174.67 &  \textbf{1722.38}&   \textbf{2126.85}   &  \textbf{2504.76}  \\ 
\bottomrule
\end{tabular*}
\begin{tablenotes}
    \item[1] Note: We take BMSE as the matrics for all these settings. AVG means the average BMSE of 20 time steps predictions. A lower value means better prediction performance. We mark the best result within a specific setting with \textbf{bold face}.
\end{tablenotes}
\end{threeparttable}
\end{table*}
\baselineskip=18pt plus.2pt minus.2pt
\parskip=0pt plus.2pt minus0.2pt

Table~\ref{Tbl:bmse} shows the comparisons of different approaches for BMSE score of the average of all 20 time steps (up to 2\;hours), 30\;minutes, 60\;minutes, 90\;minutes and 2\;hours ahead precipitation on both datasets. We can observe that FDNet outperforms the compared models on the average BMSE of 20-time-step predictions. Besides, though FDNet is not the best one for short-term prediction (30\;minutes), it shows a stable superiority for long-term predictions.

\setcounter{table}{4}
\tabcolsep 5pt
\renewcommand\arraystretch{1.3}
\begin{table*}[!htb]
\centering
\resizebox{0.95\textwidth}{!}{
\begin{threeparttable} 
\caption{\label{Tbl:ofall} Performance Comparisons of Different Approaches under CSI, HSS, BMSE and BMAE on SRAD2020 Dataset}\vspace{-2mm}
\footnotesize
\begin{tabular*}{0.93\textwidth}{ccccccccccc}
\toprule
   \multirow{2}{*}{Model}  & \multicolumn{4}{c}{CSI$\uparrow$} & \multicolumn{4}{c}{HSS$\uparrow$} & \multirow{2}{*}{BMSE$\downarrow$} & \multirow{2}{*}{BMAE$\downarrow$} \cr
  \cmidrule(lr){2-5} \cmidrule(lr){6-9} 
      & $x \ge 20 $ & $x \ge 30 $    & $x \ge 40 $ & $x \ge 50$ & $x \ge 20$ &  $x \ge 30 $ & $x \ge 40 $    & $x \ge 50 $  \cr  
    \midrule
ConvLSTM~\cite{6}     & \textbf{0.6048} & 0.4561 & 0.2186 & 0.0118 & \textbf{0.6665} & 0.5563 & 0.3333 & 0.0217 & 1753 & 8659  \\ 
TrajGRU~\cite{7}      & 0.5782 & 0.4393 & 0.2078 & 0.0257  & 0.6461 & 0.5526 & 0.3212 & 0.0473 & 1721 & \textbf{8540} \\
PredRNN~\cite{8}     & 0.5955  & 0.4507 & 0.2172  & 0.0342  & 0.6525 & 0.5542 & 0.3292 & 0.0614 & 1730 & 8769 \\
MIM~\cite{19}       & 0.5843 & 0.4512 & 0.2216  & 0.0464 & 0.6359 & 0.5545  & 0.3347 & 0.0839 & 1835 & 9151 \\
Conv-TT-LSTM~\cite{28} & 0.5762 & 0.4282&	0.2032&	0.0107	&0.6384	&0.5434	&0.2713	&0.0200	&1920  &9016\\
\textbf{FDNet}     &  0.5977	&\textbf{0.4577}	&\textbf{0.2260}	&\textbf{0.0509}	&0.6603	&\textbf{0.5665}	&\textbf{0.3415}	&\textbf{0.0910}	&\textbf{1671}  &8698\\ 
\bottomrule
\end{tabular*}
\begin{tablenotes}
    \item[1] Note: each cell contains the mean score of the 20 predicted frames. `$\uparrow$' means that the higher the score, the better. `$\downarrow$' means that the lower the score, the better. `$x \ge \tau $' means the skill score at the $\tau$\;dBZ. We mark the best result within a specific setting with \textbf{bold face}.
   \end{tablenotes}
\end{threeparttable}
}
\end{table*}
\baselineskip=18pt plus.2pt minus.2pt
\parskip=0pt plus.2pt minus0.2pt

\vspace{1mm}

\setcounter{figure}{5}
\begin{figure*}[!htb]
\centering
  \subfigure[]{
    \includegraphics[width=0.32\textwidth]{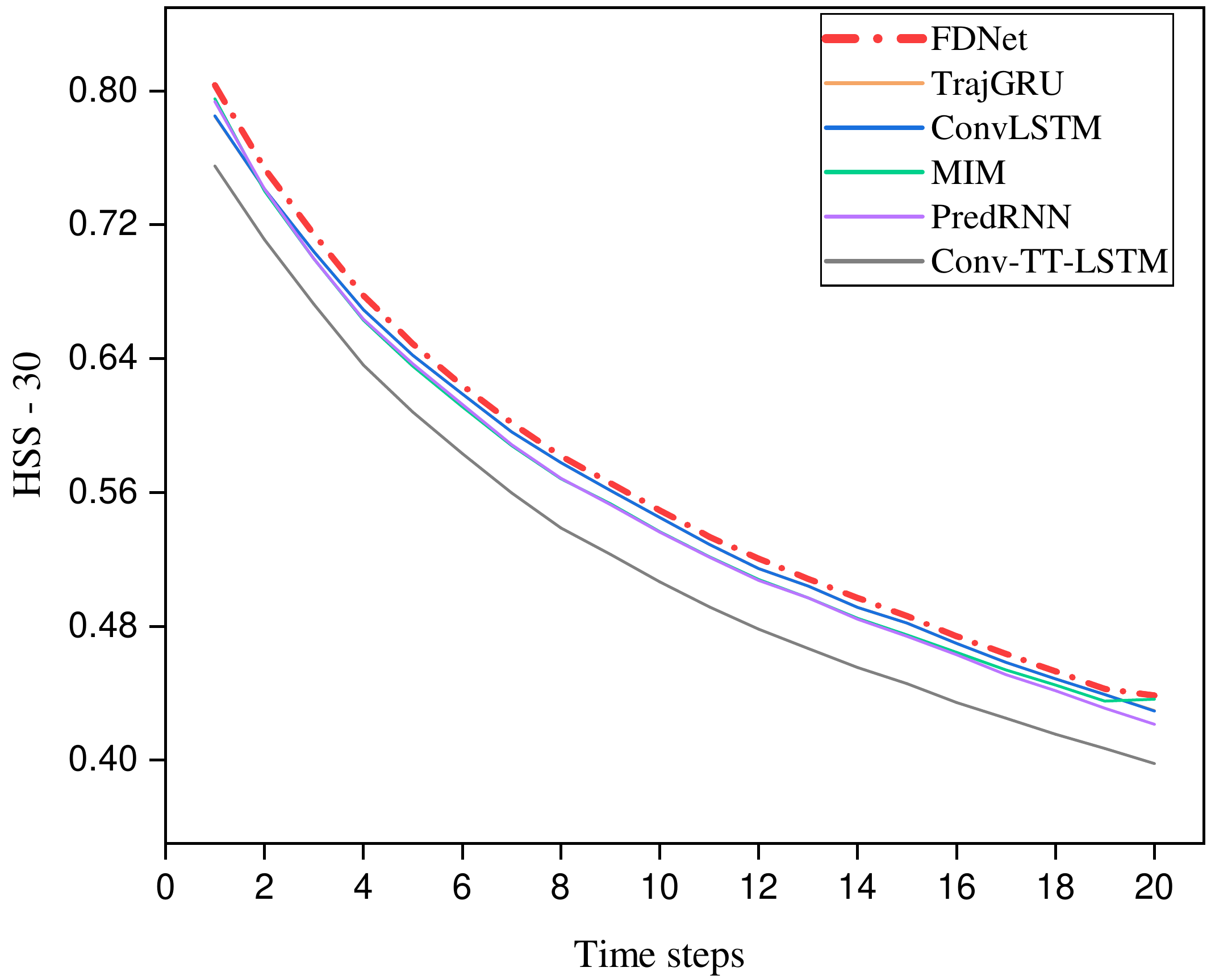}}
  \subfigure[]{
    \includegraphics[width=0.32\textwidth]{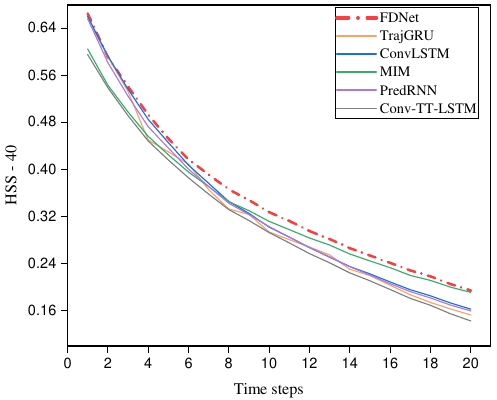}}
  \subfigure[]{
    \includegraphics[width=0.32\textwidth]{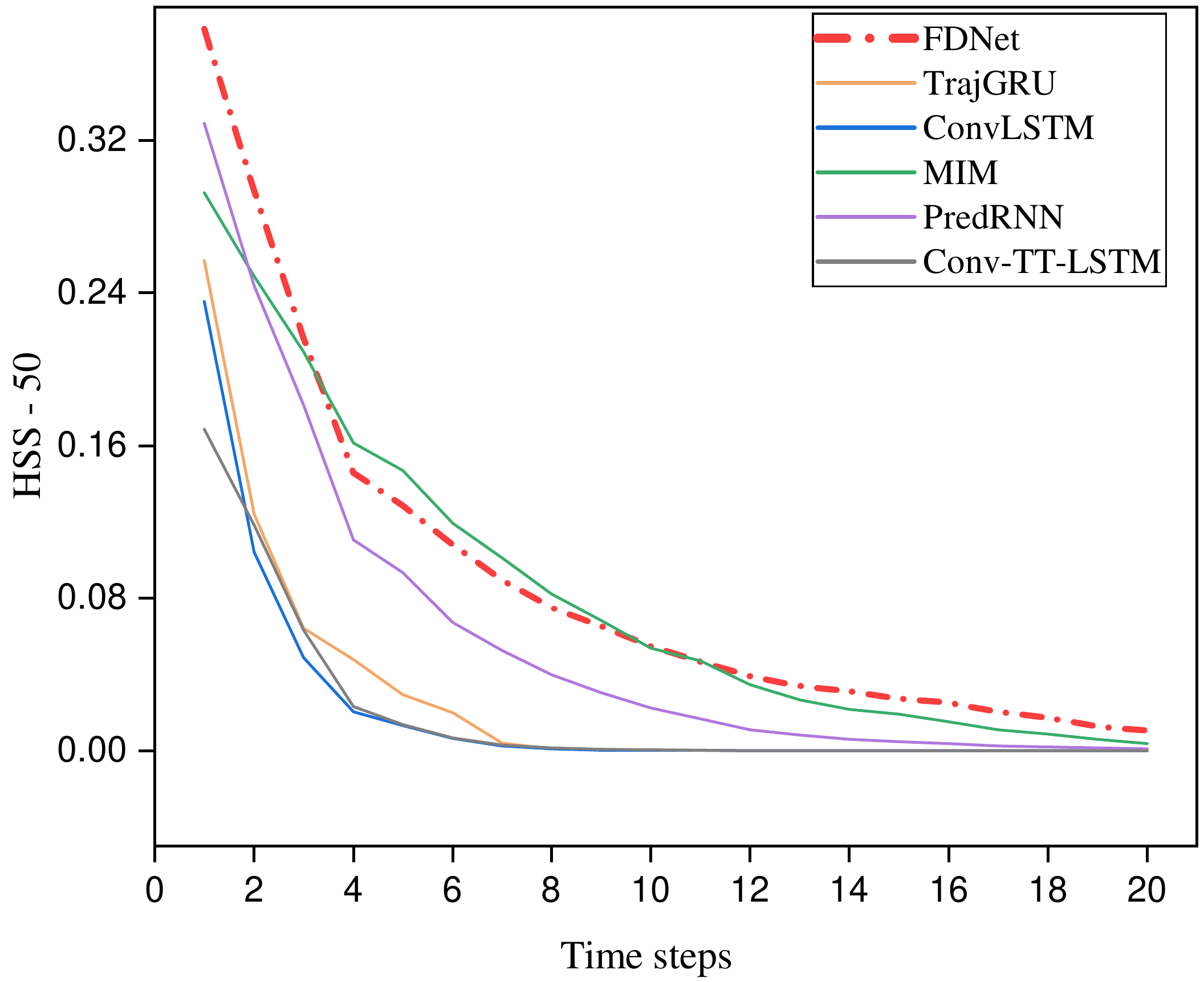}}
  \caption{Frame-wise comparison in HSS score on the SRAD2020 dataset. Higher curves indicate better forecasting results. Our FDNet performs better than baselines in most cases (except for MIM under HSS-50 at time steps 4 to 9). (a)HSS-30.(b)HSS-40.(c)HSS-50.
}
\label{fig:hss}
\end{figure*}
\baselineskip=18pt plus.2pt minus.2pt
\parskip=0pt plus.2pt minus0.2pt

Table~\ref{Tbl:ofall} shows the comparisons of different methods under CSI, HSS, BMSE and BMAE on the SRAD2020 dataset. It is worth noting that the CSI and HSS scores for thresholds 40\;dBZ and 50\;dBZ reflect whether the model's prediction of heavy precipitation is accurate and therefore it is often more concerned by meteorological experts. FDNet performs the best under CSI and HSS score for thresholds 30\;dBZ, 40\;dBZ and 50\;dBZ. We also give a frame-wise comparison of HSS score under thresholds 30\;dBZ, 40\;dBZ and 50\;dBZ in Fig.~\ref{fig:hss}. HSS-40 and HSS-50 indicate the probabilities of severe weather conditions. FDNet performs better than baseline methods in most cases (except for MIM under HSS-50 at time steps 4 to 9).

\subsection{Ablation studies}
We perform four ablation studies to analyze the respective contribution of each module in our model.

\setcounter{table}{5}
\tabcolsep 5pt
\renewcommand\arraystretch{1.3}
\begin{table*}[!]
\centering
\begin{threeparttable} 
\caption{\label{Tbl:ablation} Experimental Results under Ablated Settings}\vspace{-2mm}
\footnotesize
\begin{tabular*}{0.9\textwidth}{cccccccccc}
\toprule
 group&  seperating encoder  & no.flow & no.defor & out.flow & out.defor & BMSE & HSS-30 &HSS-40 &HSS-50 \cr
    \midrule
\multirow{2}{*}{1} & \ding{55}   & 1 & 1 &  \ding{51} & \ding{51} & 2323&	0.49&	0.11&	0.009  \\ 
&\ding{51}   & 1 & 1 & \ding{51} & \ding{51}  & 1864&	0.55	&0.33&	0.07 \\
 \midrule
\multirow{3}{*}{2} & \ding{51}  & 0 & 1 &  \ding{51} & \ding{51} & 1898&	0.55&	0.33&	0.07  \\ 
&\ding{51}   & 1 & 1 & \ding{51} & \ding{51}  & 1864&	0.55	&0.33&	0.07 \\
&\ding{51}   & 2 & 1 & \ding{51} & \ding{51}  & 1863&	0.55	&0.33&	0.06 \\
\midrule
\multirow{3}{*}{3} & \ding{51}  & 1 & 0 &  \ding{51} & \ding{51} &2245	&0.47&	0.17&	0.05 \\ 
&\ding{51}   & 1 & 1 & \ding{51} & \ding{51}  & 1864&	0.55	&0.33&	0.07 \\
&\ding{51}   & 1 & 2 & \ding{51} & \ding{51}  & 1671&	0.56&	0.34&	0.09 \\
&\ding{51}   & 1 & 3 & \ding{51} & \ding{51}  & 1934&	0.54&	0.33&	0.07 \\
\midrule
\multirow{4}{*}{4} & \ding{51}  & 1 & \ding{55} &  \ding{51} & \ding{55} &2587&	0.45&	0.15&	0.01 \\ 
&\ding{51}   & 1 & 2 & \ding{51} & \ding{55}  & 1690&	0.55&	0.33&	0.06\\
&\ding{51}   & 1 & 2 & \ding{55} & \ding{51}  & 1903&	0.53&	0.30&	0.15\\
&\ding{51}   & 1 & 2 & \ding{51} & \ding{51}  & 1671&	0.56&	0.34&	0.09 \\
\bottomrule
\end{tabular*}
\begin{tablenotes}
    \item[1] Note: We use BMSE, HSS-30, HSS-40 and HSS-50 to measure the prediction quality. We conducted four groups of comparisons. In this table, ``no.flow" means the number of ConvLSTM layers in the flow encoder; ``no.defor" means the number of ConvLSTM layers in the deformation encoder, and ``\ding{55}" in ``no.defor" means that we reduce the deformation modeling branch; ``out.flow" denotes if the output from the flow modeling branch is delivered to the decoder, and ``out.defor" denotes if the output from the deformation modeling branch is delivered to the decoder. 
   \end{tablenotes}
\end{threeparttable}
\end{table*}
\baselineskip=18pt plus.2pt minus.2pt
\parskip=0pt plus.2pt minus0.2pt

\textbf{(1) The necessity of the shape encoder and the position encoder}

In order to verify the necessity of separately extracting shape features and position features of input frames, we conduct an ablation experiment in which only one encoder is used to extract features of input frames. The results of group (1) in Table~\ref{Tbl:ablation} show that without separating encoder the model performs very poorly.

\textbf{(2) The sensitivity of our model to the number of ConvLSTM layers in the flow encoder}

We evaluate models with 0/1/2 ConvLSTM layer(s) in the flow encoder in Table~\ref{Tbl:ablation} group (2). It shows that the proposed model with 1 layer ConvLSTM in the flow encoder performs the best.

\textbf{(3) The effect of the number of ConvLSTM layers in the deformation encoder}

We also test the effect of the number of ConvLSTM layers in the deformation encoder. We evaluate models with 0/1/2/3 ConvLSTM layer(s) as shown in Table~\ref{Tbl:ablation} group (3).  The result shows that 2 layers get the best score. It is a trade-off: applying too few layers leads to inadequate deformation modeling capability while the excessively deep recurrent model leads to training difficulty.

\textbf{(4) The contribution of the flow modeling branch and the deformation modeling branch}

To see how the flow modeling branch and the deformation modeling branch contribute to the final result, we conduct a series of experiments, by 1) removing the deformation modeling branch from the proposed model (omitting the orange branch in Fig.~\ref{fig:detail}(b)); 2) only delivering the output from the flow modeling branch ($\bm{w}_t$ in Fig.~\ref{fig:detail}(b)) to the decoder; 3) feeding the concatenation of the output from the deformation modeling branch and the shape encoder to the decoder (concatenation of $\bm{d}_{t+1}$ and $\bm{s}_t$ in Fig.~\ref{fig:detail}(b)). The difference between setting (1) and setting (2) is that the latter still retains the deformation modeling information to iterate the prediction of the next step. The corresponding results are shown in Table~\ref{Tbl:ablation} group (4).  The poorly performance of the setting without the deformation encoder indicates that the deformation modeling branch is very important to the proposed model. The behavior of the restruction from the flow branch is the most similar to the proposed model, but there still remains a little gap in large rainstorm conditions ($\textgreater$ 50\;dBZ). The advantage is that the deformation branch tends to forecast heavier rainfall. Therefore the concatenation of these two complementary branches is necessary and achieves better result.

\subsection{Results Visualization}
\vspace{1mm}
\setcounter{figure}{6}
\begin{figure*}[!htb]
\centering
    \includegraphics[width=1.05\textwidth]{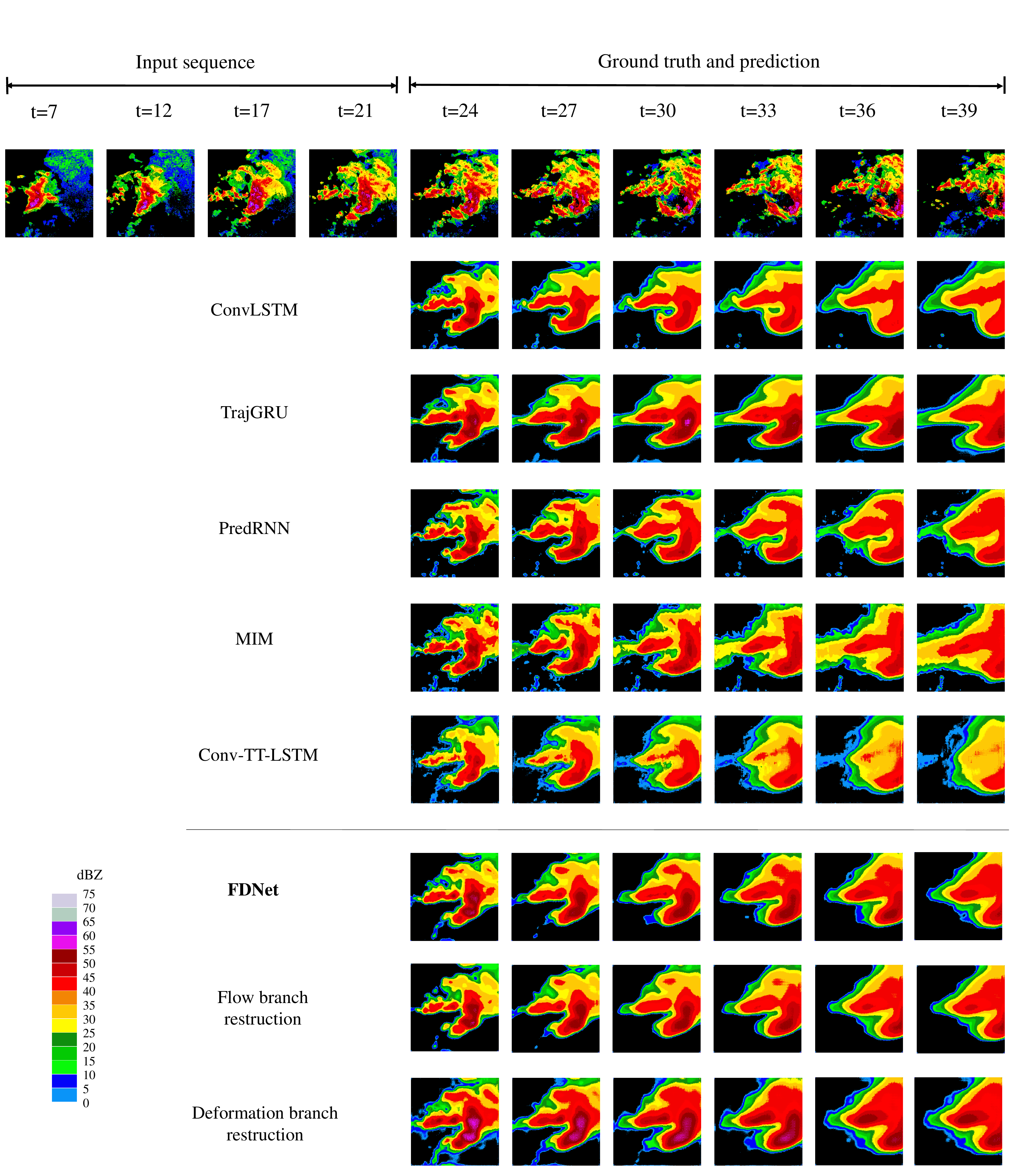}
  \caption{Showcase 1: An example of overall movement of radar echoes. The pixels tend to move from left to right as a whole. Though all these models try to predict the sequence under this movement, the baseline models all suffer from a distinct blur effect with a long tail, especially the frames in MIM.}
\label{fig:visual1}
\end{figure*}
\baselineskip=18pt plus.2pt minus.2pt
\parskip=0pt plus.2pt minus0.2pt

\setcounter{figure}{7}
\begin{figure*}[!htb]
\centering
    \includegraphics[width=1.05\textwidth]{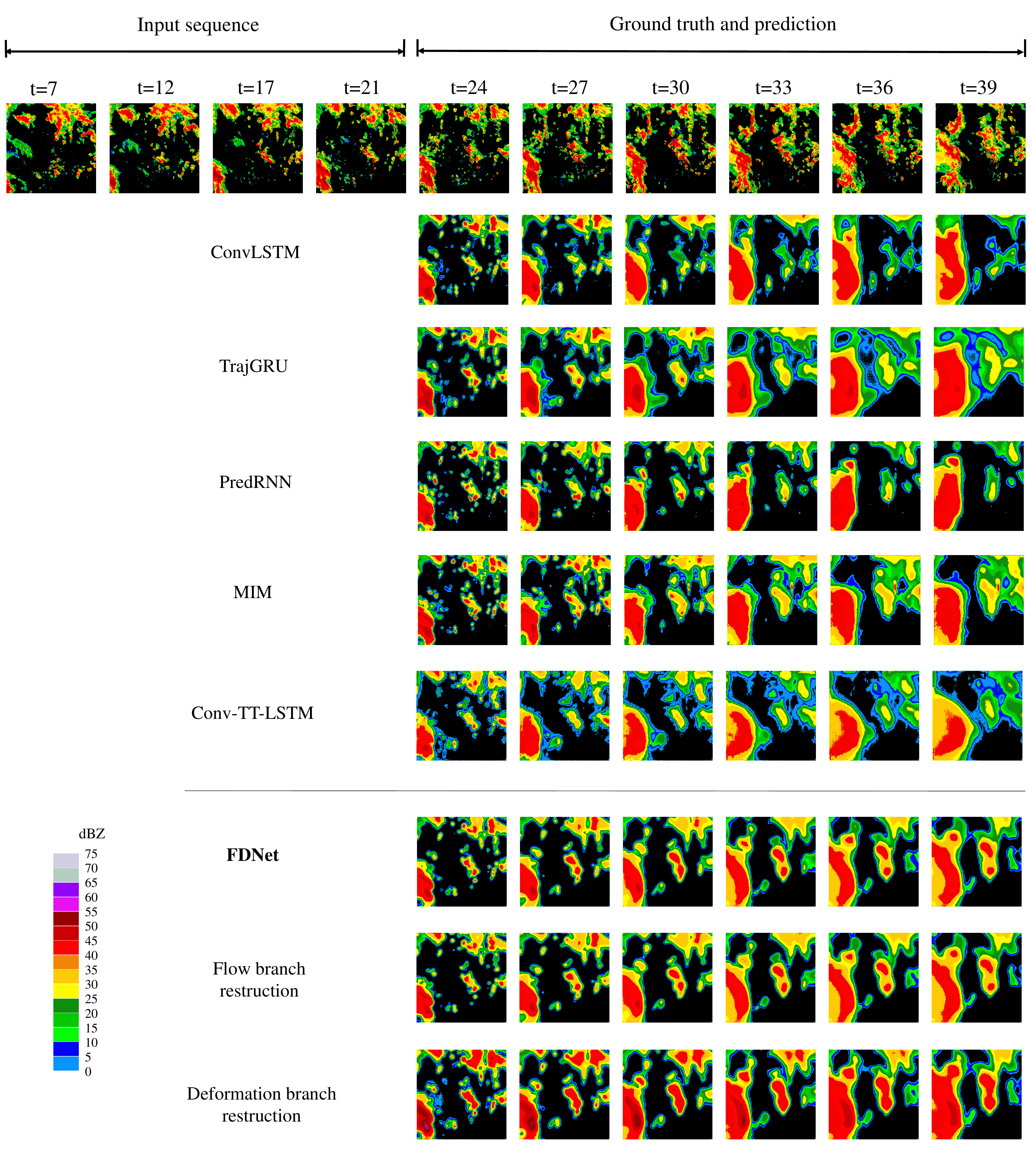}
  \caption{Showcase 2: An challenging example with accumulation and dissipation of radar echoes happening in different regions at the same time. The pixels tend to move up and right as a whole, but zooming in to the detail, the echoes at the left bottom dissipate, while the rainfall range in the middle is expanding. All baseline models fail to perceive the dissipation at the left bottom and lose the heavy rainfall information in the middle area. Only FDNet captures all these fine-grained evolutions and predicts relatively correct frames, especially for longer future time steps.
}
\label{fig:visual2}
\end{figure*}
\baselineskip=18pt plus.2pt minus.2pt
\parskip=0pt plus.2pt minus0.2pt

\setcounter{figure}{8}
\begin{figure*}[!htb]
\centering
    \includegraphics[width=1.05\textwidth]{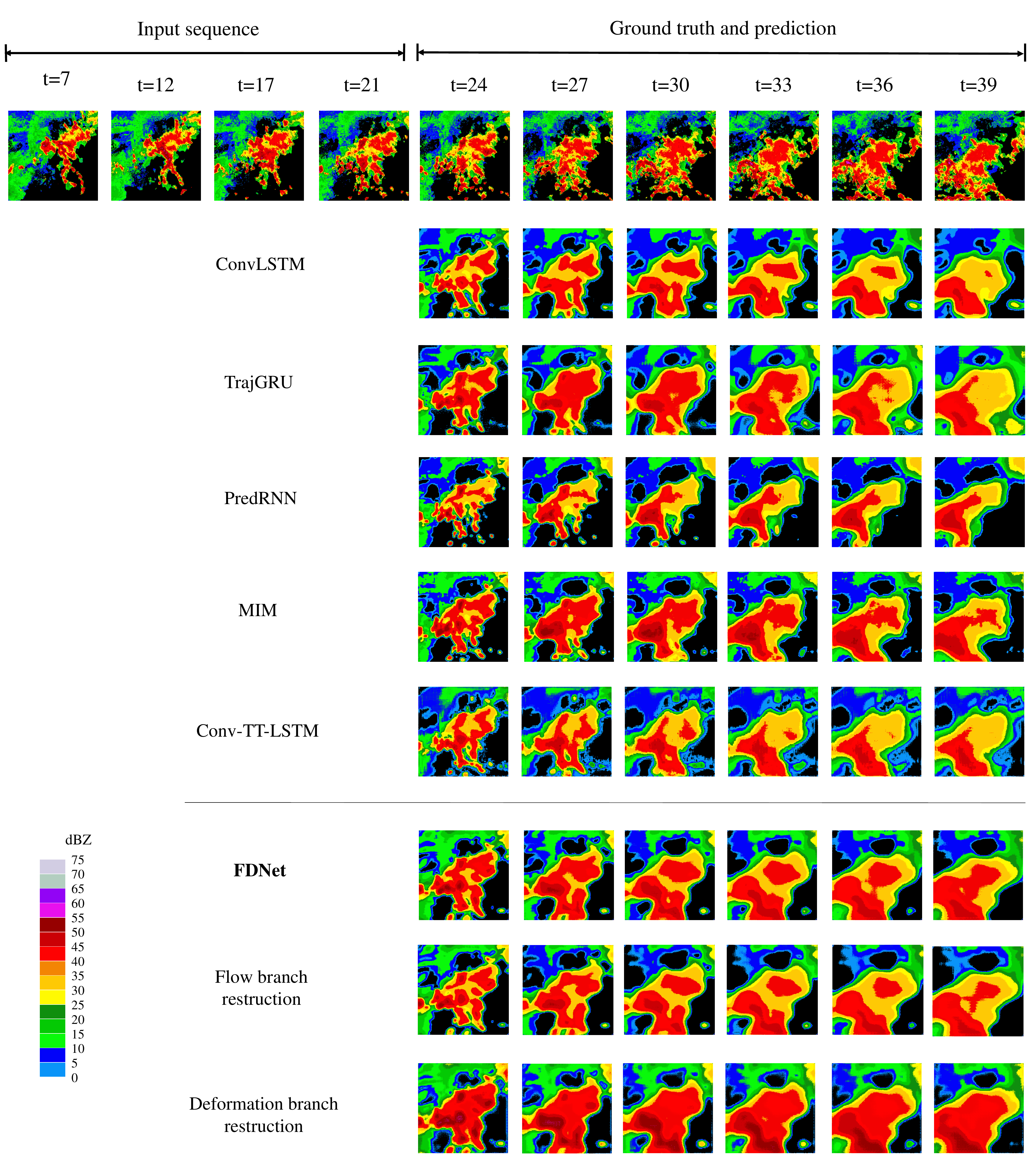}
  \caption{Showcase 3: The intensities of radar echoes remain the same on the whole, with a slight increase. All other methods predict that the rainfall in the middle area would get abating. Only FDNet successfully predicts the long-lasting heavy rain.
}
\label{fig:visual3}
\end{figure*}
\baselineskip=18pt plus.2pt minus.2pt
\parskip=0pt plus.2pt minus0.2pt

\setcounter{figure}{9}
\begin{figure*}[!htb]
\centering
    \includegraphics[width=1.05\textwidth]{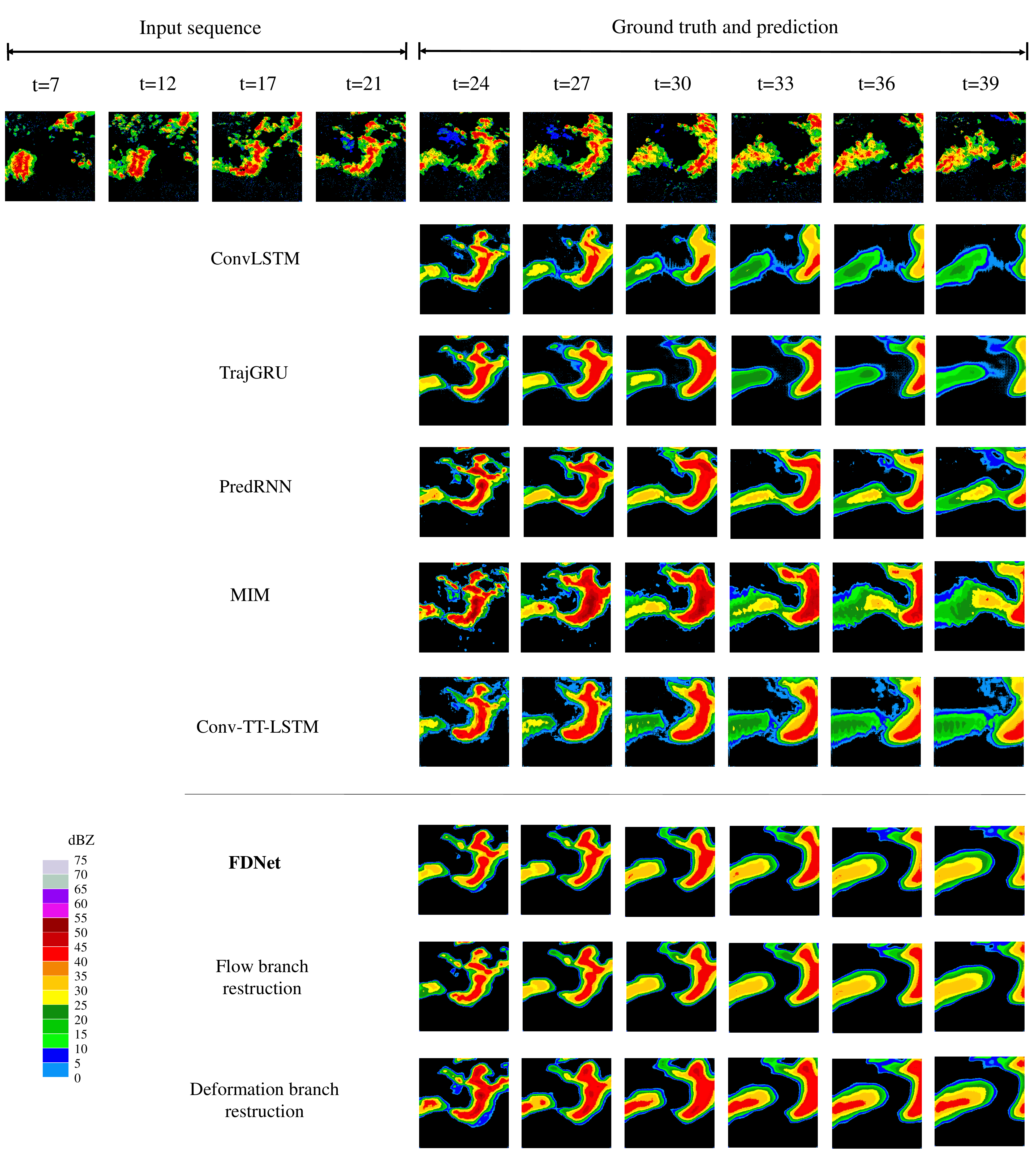}
  \caption{Showcase 4: An example of new coming rainfall. The deformation modeling is particularly sensitive to newly generated rainfall. For example, the restruction result from the deformation branch successfully predicts the new coming heavy rainfall at the left bottom, which makes FDNet's predictions much closer to the ground truth numerically than other models.
}
\label{fig:visual4}
\end{figure*}
\baselineskip=18pt plus.2pt minus.2pt
\parskip=0pt plus.2pt minus0.2pt

We visualize four typical sequences of predicted radar echo map in Fig.\ref{fig:visual1}, Fig.\ref{fig:visual2}, Fig.\ref{fig:visual3} and Fig.\ref{fig:visual4}, respectively. The first line shows the original radar echo map sequence. We compare the predicted results of our model with five baseline methods. In order to see how our proposed two pathways work, we also give the visualizations of partial predictions labelled with ``flow branch restruction" and ``deformation branch restruction". This is done by delivering the output from the flow encoder($\bm{w}_t$ in Fig.~\ref{fig:detail}(b)) to the decoder and the concatenation of the output from the deformation encoder and the shape encoder to the decoder (concatenation of $\bm{d}_{t+1}$ and $s_t$ in Fig.~\ref{fig:detail}(b)), respectively. All compared models take 21 historical radar echo images as inputs, and predict the next 20 images (radar echo maps for the next two hours). 

It can be seen that the flow encoder mainly focuses on the movement of radar echoes as a whole, and can remember the general outline and scope. It helps the model avoid extreme expansions of radar echoes in prediction which appear in other methods (see Fig.\ref{fig:visual1} and Fig.\ref{fig:visual2}). The deformation encoder is very sensitive to pixel changes, even the information lost during `\texttt{warp}' procedure. Therefore it helps the proposed model remember the heavy rainfall information for a long time (see Fig.\ref{fig:visual2} and Fig.\ref{fig:visual3}), while other models all lose this information step by step when modeling motion and transformation together. The deformation branch also equips the model with capability of predicting new coming rainfall (see Fig.\ref{fig:visual4}). Overall, benefiting from these two separate encoding pathways, FDNet's results are not only sharp enough but also more deterministic in future predictions.

\section{Conclusions}
In this paper, we have proposed a Flow-Deformation network for precipitation nowcasting. The proposed model employs two parallel cross encoding pathways and learns to decompose optical flow field motion and morphologic deformation of radar echoes. In this model, we design a flow encoder to capture tendencies of the bodily movements of radar echoes, and invent a deformation encoder to perceive the variations of the deformation. Experimental results suggest that separate modeling of flow and deformation reduces the uncertainty of the forecast and slows down the tendency of image blurring. Our model performs favorably compared with the state-of-the-art methods on precipitation nowcasting, especially for relatively longer future predictions. For future work, we plan to train an intelligent model to distinguish different evolution patterns among radar echo sequences, and treat them with the most suitable model individually.

\label{last-page}
\end{multicols}
\label{last-page}
\end{document}